%% file: main.tex
\documentclass[conference]{IEEEtran}

\IEEEoverridecommandlockouts
\usepackage{amsmath,amssymb,amsfonts}
\usepackage{graphicx}
\graphicspath{ {./assets/} }
\usepackage{textcomp}
\usepackage{biblatex}
\usepackage{forest}
\usepackage{svg}
\usepackage{multirow}
\usepackage{tikz}
\usepackage{adjustbox}
\usepackage{subfig}
\usepackage{caption}
\usepackage{mdframed}
\usepackage{soul}
\usepackage{algorithm}
\usepackage{algpseudocode}
\usepackage{lipsum}
\usepackage{tabularx}
\usepackage{makecell}
\usepackage{float}
\usepackage{url}
\usepackage{hyperref}
\usepackage{flushend}

\newcolumntype{L}{>{\raggedright\arraybackslash}X}
\usepackage[normalem]{ulem}
\useunder{\uline}{\ul}{}
\usetikzlibrary{shapes, arrows, positioning, arrows.meta, mindmap}

\input{./tikz/shapes.tikz}

\def\BibTeX{{\rm B\kern-.05em{\sc i\kern-.025em b}\kern-.08em
    T\kern-.1667em\lower.7ex\hbox{E}\kern-.125emX}}

\DeclareRobustCommand{\IEEEauthorrefmarktwo}[1]{\smash{\textsuperscript{\footnotesize #1}}}

\begin{document}

%\title{HowkGPT: Unmasking ChatGPT's Influence in Student Homework through Context-Aware Perplexity Analysis}
\title{HowkGPT: Investigating the Detection of ChatGPT-generated University Student Homework through Context-Aware Perplexity Analysis}

% \author{
% \IEEEauthorblockN{Christoforos Vasilatos}
% % \IEEEauthorblockA{\textit{dept. name of organization (of Aff.)} \\
% \textit{New York University Abu Dhabi}\\
% Abu Dhabi, UAE \\
% cv43@nyu.edu
% \and
% \IEEEauthorblockN{Manaar Alam}
% % \IEEEauthorblockA{\textit{dept. name of organization (of Aff.)} \\
% \textit{New York University Abu Dhabi}\\
% Abu Dhabi, UAE \\
% ma6996@nyu.edu
% \and
% \IEEEauthorblockN{Talal Rahwan}
% % \IEEEauthorblockA{\textit{dept. name of organization (of Aff.)} \\
% \textit{New York University Abu Dhabi}\\
% Abu Dhabi, UAE \\
% tr72@nyu.edu
% \and
% \IEEEauthorblockN{Yasir Zaki}
% % \IEEEauthorblockA{\textit{dept. name of organization (of Aff.)} \\
% \textit{New York University Abu Dhabi}\\
% Abu Dhabi, UAE \\
% yz48@nyu.edu
% \and
% \IEEEauthorblockN{Michail (Mihalis) Maniatakos}
% % \IEEEauthorblockA{\textit{dept. name of organization (of Aff.)} \\
% \textit{New York University Abu Dhabi}\\
% Abu Dhabi, UAE \\
% mm6446@nyu.edu
% }

\author{
\IEEEauthorblockN{
    Christoforos Vasilatos\IEEEauthorrefmarktwo{1}, %\IEEEauthorrefmark{1},
    Manaar Alam\IEEEauthorrefmarktwo{1},
    Talal Rahwan\IEEEauthorrefmarktwo{2},
    Yasir Zaki\IEEEauthorrefmarktwo{2}, and
    Michail Maniatakos\IEEEauthorrefmarktwo{1}%\IEEEauthorrefmark{2}
    }
    \IEEEauthorblockA{\IEEEauthorrefmarktwo{1}Center for Cyber Security, New York University Abu Dhabi, United Arab Emirates}
    \IEEEauthorblockA{\IEEEauthorrefmarktwo{2}Division of Science, New York University Abu Dhabi, United Arab Emirates}
    % \IEEEauthorblockA{
    % Email:
    %     \IEEEauthorrefmark{1}cv43@nyu.edu,
    %     \IEEEauthorrefmark{2}mm6446@nyu.edu
    %     }
}

\maketitle

\begin{abstract}
\input{sections/0_abstract}
\end{abstract}


\section{Introduction}
\input{sections/1_introduction}

\section{Background}
\input{sections/2_background}

\section{Dataset Construction}
\label{sec:dataset_construction}
\input{sections/3_dataset_construction}

\section{Methodology}
\label{sec:methodology}
\input{sections/4_methodology}

\section{Evaluation}
\label{sec:evaluation}
\input{sections/5_evaluation}

\section{Discussion}
\label{sec:discussion}
\input{sections/6_discussion}

\section{Conclusion}
\label{sec:conclusion}
\input{sections/7_conclusion}

\section*{Resources}
HowkGPT can be found at \url{https://howkgpt.nyuad.nyu.edu/}.

\printbibliography

% \clearpage

% \appendix
% \label{sec:appendix}
% \input{sections/100_appendix}

\end{document}

%% file: tikz/shapes.tikz
\tikzstyle{window2} = [rectangle, rounded corners, 
minimum width=1cm,
minimum height=0.5cm,
text width=0.4\columnwidth,
text centered,
draw=black]

\tikzstyle{window} = [rectangle, rounded corners, 
minimum width=1cm,
minimum height=0.5cm,
text width=0.90\columnwidth,
text centered,
draw=black]

\tikzstyle{startstop} = [rectangle, rounded corners, 
minimum width=1cm,
minimum height=0.5cm,
text centered,
draw=black,
fill=red!30]

\tikzstyle{io} = [trapezium,
trapezium stretches=true,
trapezium left angle=70,
trapezium right angle=110,
minimum width=1cm,
minimum height=0.5cm,
text centered,
draw=black, fill=blue!30]

\tikzstyle{process} = [rectangle, 
minimum width=1cm,
minimum height=0.5cm,
text centered,
text width=3cm,
draw=black,
fill=orange!30]

\tikzstyle{decision} = [diamond, 
minimum width=1cm,
minimum height=0.5cm,
text centered,
draw=black,
aspect=1.9,
inner sep=3pt,
fill=green!30]

\tikzstyle{arrow} = [thick,->,>=stealth]

%% file: sections/0_abstract.tex
As the use of Large Language Models (LLMs) in text generation tasks proliferates, concerns arise over their potential to compromise academic integrity. The education sector currently tussles with distinguishing student-authored homework assignments from AI-generated ones. This paper addresses the challenge by introducing HowkGPT, designed to identify homework assignments generated by AI. HowkGPT is built upon a dataset of academic assignments and accompanying metadata~\cite{ibrahim2023perception} and employs a pretrained LLM to compute perplexity scores for student-authored and ChatGPT-generated responses. These scores then assist in establishing a threshold for discerning the origin of a submitted assignment. Given the specificity and contextual nature of academic work, HowkGPT further refines its analysis by defining category-specific thresholds derived from the metadata, enhancing the precision of the detection. This study emphasizes the critical need for effective strategies to uphold academic integrity amidst the growing influence of LLMs and provides an approach to ensuring fair and accurate grading in educational institutions.

%% file: sections/1_introduction.tex
The recent proliferation of Large Language Models (LLMs) has resulted in their widespread availability as web-based applications. These models demonstrate an impressive capability to respond to queries and interact in a manner that closely resembles human communication. Central to these LLMs are Transformer models, which present a broad spectrum of applications, including but not limited to content recommendation~\cite{liu2023pretrain}, language translation~\cite{vilar2022prompting}, sentiment analysis~\cite{yadav2020sentiment}, text classification~\cite{kant2018practical}, and, most notably, text generation~\cite{celikyilmaz2021evaluation}. Prominent among these web applications are OpenAI's ChatGPT~\cite{openai-chatgpt}, built on the GPT-4 architecture~\cite{GPT4} and Google Bard~\cite{google-bard} as well as Google AI Test Kitchen~\cite{AI-Test-Kitchen}, built on the LaMDA Transformer~\cite{LaMDA, adiwardana2020towards}. These tools have gained substantial attention within scholarly communities and educational institutions. However, the emergence of such potent tools, particularly in their potential to simplify homework completion, poses an intriguing challenge regarding fair and accurate evaluation and grading of student homework.

The advancement of LLMs mentioned above has brought unprecedented capabilities in generating human-like text for academic assignments and programming tasks~\cite{khalil2023chatgpt}. This evolution has necessitated the development of efficient mechanisms to distinguish student-authored submissions from those generated by LLMs. This need is rooted in the fundamental principle of academic integrity, ensuring impartial evaluation and promoting an environment conducive to authentic learning. The absence of such a mechanism places educators at risk of incorrectly grading AI-generated work, thereby distorting the fairness in academic grading. Furthermore, reliance on AI-generated homework may impede students from understanding their coursework deeply, consequently undermining the educational experience. Recent research has primarily concentrated on distinguishing between AI-generated and human-written text within a general context~\cite{mitrović2023chatgpt,ippolito2020automatic,tang2023science,mitchell2023detectgpt,munyer2023deeptextmark,li2023origin,sadasivan2023aigenerated}. However, identifying AI-generated homework assignments presents unique challenges compared to identifying general AI-generated content. One of the key reasons is the specificity and contextuality associated with academic assignments. These assignments often require the application of specific theories, principles, and problem-solving skills. While AI-generated general content might exhibit noticeable inconsistencies in the broader context, the narrower and more structured scope of academic assignments may mask such anomalies, making the detection process more complex. Hence, distinguishing AI-generated homework requires more refined and context-aware algorithms.

In this study, we introduce a tool called HowkGPT designed to evaluate whether academic assignments are generated by ChatGPT or written independently by students. To begin, we use a dataset composed of academic assignments developed by Ibrahim et al.~\cite{ibrahim2023perception}. This dataset includes accompanying metadata, which provides multiple categorizations for the dataset. HowkGPT utilizes the dataset and its associated metadata to compute the perplexity metric for responses submitted by students and ChatGPT. It should be noted that the computation of perplexity values for an LLM requires its white-box access. As a result, given the current inaccessibility of LLMs presently utilized by ChatGPT (i.e., GPT-3.5 and GPT-4), we resort to the utilization of a pretrained GPT-2 model~\cite{gpt2-huggingface, radford2019language}. The principal objective of HowkGPT is to precisely define a threshold, using the perplexity score in conjunction with metadata, to identify the origin of an academic assignment correctly. We demonstrate that categorizing academic assignments and having category-wise thresholds facilitates better accuracy than calculating a single threshold value across the entire dataset without metadata categorization.

\subsection*{\textbf{Our Contributions:}}
\begin{enumerate}
    \item We propose a novel multi-level approach to detect AI-generated text focusing on university student homework. Our method utilizes metadata categorization from an academic dataset to enhance the perplexity metric used to detect whether a given assignment has been student-authored or AI-generated.

    \item We perform and present extensive experiments to evaluate accuracy of detection using knowledge and cognitive dimensions.

    \item We develop a publicly available web application: \url{https://howkgpt.hpc.nyu.edu/}. This experimental platform performs real-time assessments on assignment submissions.
\end{enumerate}

%% file: sections/2_background.tex
\subsection{ChatGPT}
ChatGPT, a state-of-the-art language model developed by OpenAI, leverages the GPT-4~\cite{openai-blog} architecture for its paid version and GPT-3.5 for the free version. This language model is adept at generating human-like text, responding to questions, and providing recommendations across various contexts, including mathematics, programming, and numerous other knowledge domains. Building upon the success of its predecessors, GPT-3 and GPT-2, ChatGPT integrates a larger dataset, advanced training methodologies, and an improved transformer architecture. The dataset comprises many sources, such as books, articles, and websites, ensuring a comprehensive understanding of language and knowledge representation. ChatGPT continually evolves through user interaction and ongoing enhancements introduced by the development team. Its advanced capabilities have resulted in its widespread adoption across diverse applications, such as content generation, virtual assistants, and customer support. However, the ethical implications and potential misuse of such potent language models remain a pressing concern for researchers and developers alike.

\subsection{Perplexity}\label{subsec:backround:ppl}
Perplexity is a statistical metric used in Natural Language Processing and, more specifically, in language models. It measures how well a probability model predicts a sample and is used to compare the performance of different models on the same dataset. The concept of perplexity for language models originated from the field of information theory. In information theory, perplexity measures how uncertain a prediction model is, given the actual outcome. When applied to language models, this concept is adapted to estimate the average uncertainty of predicting the next word in a sequence given the previous words. A lower perplexity score indicates that the language model is better at predicting the sample. This is because a lower perplexity means the model is less uncertain about its predictions.

Perplexity, in a more precise definition, is characterized as the exponentiated average negative log-likelihood of a sequence. Here, a sequence is denoted to be an ordered list of words, or in our specific context, we can consider these words as tokens. Let us assume a tokenized sequence of length $t$ as
\begin{center}
    $X = \{x_0, x_1, \dots, x_t\}$
\end{center}

Then, the perplexity of the sequence $X$ can be mathematically expressed through the following function, denoted as $PPL$:

\begin{equation}\label{eq:ppl}
PPL(X) = exp\Biggl\{-\frac{1}{t}\sum_{i}^{t} logp_\theta (x_i|x_{<i})\Biggl\}
\end{equation}

with $logp_\theta (x_i|x_{<i})$ that can be rewritten as $logP(X_i|\theta)$, where $P(X_i|\theta)$ is the probability (or likelihood) of obtaining the data point $X_i$ given the parameter values $\theta$. More specifically, in our case, $logp_\theta (x_i|x_{<i})$ is the log-likelihood of the i-th token conditioned on the preceding tokens $x_{<i}$, given $\theta$ which represents the parameter values of the model or else the values of the tokens in a given context.
 
Perplexity is frequently employed in practice as a comparative measure for evaluating the performance of various language models on specific tasks. This concept has been repurposed by researchers who developed the GLTR tool~\cite{gehrmann2019gltr} to determine whether an AI has generated a particular piece of text. More precisely, the perplexity of a given text can act as an indirect indicator of its association with a language model. This relationship is established on the premise that a lower perplexity score implies a higher likelihood of the text being generated by a language model.

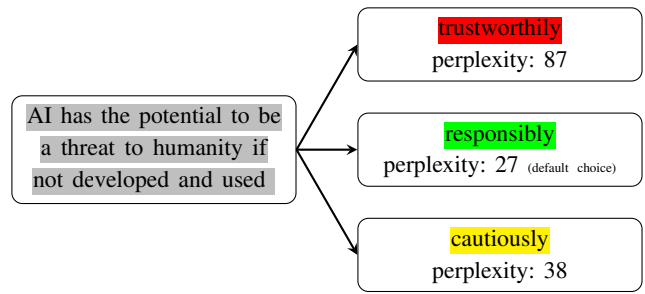
\begin{figure}[!b]
\centering
\begin{tikzpicture}[node distance = 0.6cm]
\sethlcolor{lightgray}

\hl{AI has the potential to be a threat to humanity if not developed and used}

\node (window1) [window2] {\small{\hl{AI has the potential to be a threat to humanity if not developed and used }}};
\node (window3) [window2, right of=window1, xshift=4cm] {\small{\sethlcolor{green}\hl{responsibly}\\perplexity: 27 \tiny{(default choice)}}};
\node (window2) [window2, above of=window3, yshift=-2cm] {\small{\sethlcolor{yellow}\hl{cautiously}\\perplexity: 38}};
\node (window4) [window2, below of=window3, yshift=2cm] {\small{\sethlcolor{red}\hl{trustworthily}\\perplexity: 87}};

\draw [arrow] (window1.east) -- (window2.west);
\draw [arrow] (window1.east) -- (window3.west);
\draw [arrow] (window1.east) -- (window4.west);

\end{tikzpicture}
\caption{An illustrative example of perplexity scores computed using HowkGPT for different options as the next word given a specific context (highlighted in gray), where `responsibly' is the default choice of ChatGPT.}
\label{fig:ppl_choices}
\end{figure}

Figure~\ref{fig:ppl_choices} shows an illustration, providing insight into the computation of perplexity scores for the next word or token generated by a language model given a specific context (highlighted in gray). The figure shows three potential choices for the next word: `trustworthily', `responsibly', and `cautiously'. For this instance, the default selection of the ChatGPT model is `responsibly'. For the sake of this demonstration, we select two synonyms of `responsibly' as `trustworthily' and `cautiously'. Utilizing the proposed HowkGPT, we compute the perplexity score for each option. The results depicted in the figure demonstrate that `responsibly' yields the lowest perplexity score. Conversely, `trustworthily' is an improbable selection, causing an overall increase in the perplexity score by $3.2\times$ compared to the baseline sentence with `responsibly'.

% The two alternative choices which are synonyms of the baseline indicate that especially, \textit{trustworthily}, is a very impossible word to be selected and thus the perplexity of the sentence becomes 3.2 times bigger than the baseline sentence.

%% file: sections/3_dataset_construction.tex
The dataset used in our study is the output of an academic survey performed by Ibrahim et al.~\cite{ibrahim2023perception}. The survey consists of 10 different questions from each of the thirty two selected courses offered at New York University Abu Dhabi (NYUAD). The following are examples of some of these courses:
\begin{enumerate}
    \item Data Structures
    \item Introduction to Public Policy
    \item Quantitative Synthetic Biology
    \item Cyberwarfare
    \item Object Oriented Programming
    \item Structure and Properties of Civil Engineering Materials
    \item Biopsychology
    \item Climate/Change
    \item Management and Organizations
\end{enumerate}

The courses were explicitly selected by a diverse group of thirty-one faculty members from NYUAD. Faculty members belonging to various domains, such as Computer Science, Political Science, Mathematics, etc., initially integrated these questions into their course homework assignments. The responses to the survey consist of three randomly selected student replies for each question. Simultaneously, the same questions were presented to the OpenAI ChatGPT web application~\cite{openai-chatgpt-api} across various sessions. This process yielded three unique AI-generated responses for each question, further augmenting the dataset.

The research performed in this paper strictly adhered to all pertinent guidelines and regulations. The consent of all 398 participants was duly procured, ensuring their informed agreement in every phase of the study. Importantly, the dataset creation procedure received approval from the Institutional Review Board of New York University Abu Dhabi under the approval code HRPP-397 2023-5.

\begin{figure}[!t]
\centering
\resizebox{\columnwidth}{!}{%
\begin{tikzpicture}[scale=1.3, mindmap, grow cyclic, every node/.style={scale=1.3, concept}, concept color=orange!40,
level 1/.append style={level distance=2.6cm, sibling angle=120},
level 2/.append style={level distance=2.6cm, sibling angle=45}]

\node[scale=0.6]{Question Categorization}
child [concept color=blue!30] { node {Knowledge Dimension}
	child { node {Conceptual}}
	child { node {Factual}}
	child { node {Procedural}}
	child { node {Metacognitive}}
}
child [concept color=yellow!30] { node {Cognitive Process Dimension}
	child { node {Apply}}
	child { node {Understand}}
	child { node {Analyze}}
	child { node {Evaluate}}
	child { node {Create}}
 	child { node {Remember}}
}
child [concept color=teal!40] { node {Include}
	child { node {Math}}
	child { node {Code}}
        child { node {Author Book}}
        child { node {Trick}}
};
\end{tikzpicture}%
}
\caption{The categorization defined by the professors providing the questions.}
\label{fig:question_categorization}
\end{figure}
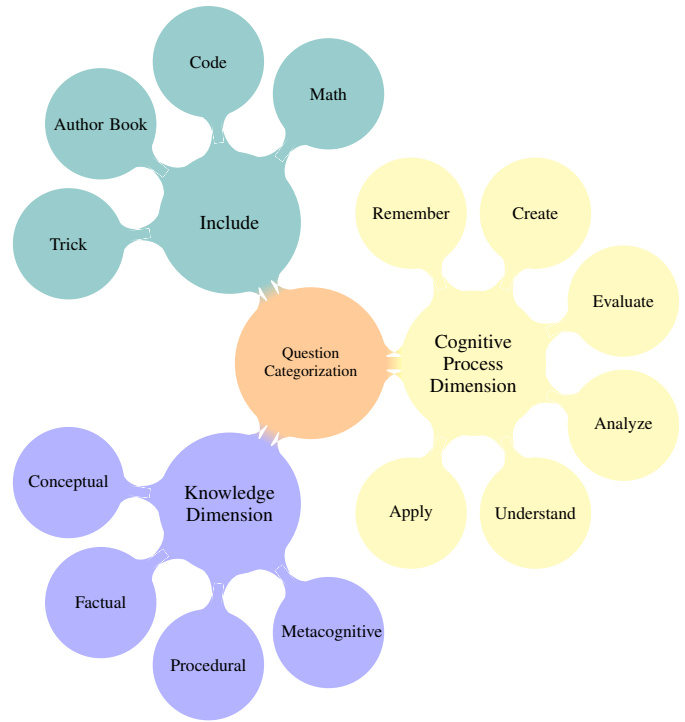

Each faculty member was also responsible for providing supplementary metadata related to the questions. The metadata includes the categorization of the questions according to various parameters, including the knowledge dimension, cognitive process dimensions, and the inclusion of specific attributes within the responses. A concise summary of the entire question categorization is shown in Figure~\ref{fig:question_categorization}.
The detailed descriptions of these categorizations, along with their corresponding subcategories, are presented in Table~\ref{tab:dimension_subcategory_explanation}.
\begin{table*}[!t]
    \centering
    \caption{Explanation of question categorization and subcategories (as shown in Figure~\ref{fig:question_categorization}) derived from~\cite{ibrahim2023perception}.}
    \begin{tabularx}{\textwidth}{ | L | L || L | L | L || L | L |}
        \hline
        \multicolumn{2}{| c ||}{\textbf{Knowledge Dimension}} & \multicolumn{3}{ c ||}{\textbf{Cognitive Process Dimension}} & \multicolumn{2}{ c |}{\textbf{Include Dimension}}\\
        \hline
        \textbf{Conceptual} & \textbf{Factual} & \textbf{Remember} & \textbf{Understand} & \textbf{Apply} & \textbf{Code} & \textbf{Math}\\
        \hline
        \scriptsize{The interrelationships among the basic elements within a larger structure that enable them to function together.} & 
        \scriptsize{The basic elements that students must know to be acquainted with a discipline or solve problems in it.} &
        \scriptsize{Retrieving relevant knowledge from long-term memory.} & 
        \scriptsize{Determining the meaning of instructional messages, including oral, written, and graphic communication.} & 
        \scriptsize{Carrying out or using a procedure in a given situation.}&
        \scriptsize{involves mathematics}&
        \scriptsize{involves code snippets}\\
        \hline
        \textbf{Procedural} & \textbf{Metacognitive} & \textbf{Analyze} & \textbf{Evaluate} & \textbf{Create} & \textbf{Author Book} & \textbf{Trick}\\
        \hline
        \scriptsize{How to do something; methods of inquiry, and criteria for using skills, algorithms, techniques, and methods.} &
        \scriptsize{Knowledge of cognition in general as well as awareness and knowledge of one’s own cognition.} &
        \scriptsize{Breaking material into its constituent parts and detecting how the parts relate to one another and to an overall structure or purpose.} &
        \scriptsize{Making judgments based on criteria and standards.} &
        \scriptsize{Putting elements together to form a novel, coherent whole or make an original product.}&
        \scriptsize{Requires knowledge of a specific author, paper/book, or a particular technique/method}&
        \scriptsize{A trick question is a question that is designed to be difficult to answer or understand, often with the intention of confusing or misleading the person being asked.}\\
        \hline
    \end{tabularx}
    \label{tab:dimension_subcategory_explanation}
\end{table*}
The categorization framework is derived from Anderson and Krathwohl's taxonomy~\cite{krathwohl2002revision}. Each question can be uniquely identified by a single subcategory within the knowledge dimension, but it may also align with multiple subcategories under the cognitive process dimension. Moreover, the attributes of `trick', `author book', `code', and `math' exhibit binary characteristics, indicating that a response may require the inclusion or exclusion of any combination of these elements.

The metadata discussed above serves to differentiate the texts into diverse categories. While LLMs exhibit robust performance in certain domains, they display deficiencies in others, necessitating fine-tuning to enhance performance in specific tasks or domains~\cite{gururangan2020dont}. The metadata-driven categorization is anticipated to provide a foundation for an additional layer of properties that can discriminate the output of an LLM and the human-written text. The proposed methodology in Section~\ref{sec:methodology} applies broadly and does not limit itself to a specific set of texts or knowledge domains. Furthermore, provided the diversity of the dataset, it is ensured that the method effectively covers an extensive array of possible textual domains.

%% file: sections/4_methodology.tex
\subsection{Motivation and Overview}
As discussed in Section~\ref{subsec:backround:ppl}, perplexity serves as a metric for evaluating the performance of a language model. In addition, it can also be leveraged to ascertain the likelihood of whether a set of tokens (or words) chosen from a specific segment of a text have a high probability of being generated by an LLM. Hashimoto et al.~\cite{hashimoto2019unifying} demonstrated the potential of discerning between human-written and AI-generated texts based on model likelihood. Perplexity, interchangeably referred to as predictive likelihood, is fundamentally the exponentiated average of negative log-likelihood. Hence, it can be deduced that high perplexity values indicate a lower probability for an LLM to generate the particular tokens within the text. On the contrary, low perplexity values indicate a higher likelihood of the tokens being generated by the LLM model. This inference stems from the definition of perplexity as provided in Equation~\ref{eq:ppl} since the probability is inversely proportional to the exponent of perplexity.

In order to compute perplexity, we resort to a pre-trained GPT2 model~\cite{gpt2-huggingface, radford2019language} due to the unavailability of models that ChatGPT currently deploys (i.e., GPT-3.5 and GPT-4). Nevertheless, the perplexity scores computed for texts generated by ChatGPT are comparatively low, implying a similarity in the functionality between the two models. The preliminary objective is to develop a tool based on the dataset specified in Section~\ref{sec:dataset_construction}, concentrating on texts generated to answer homework questions. This constraint enhances the precision of the tool's predictions, given that the dataset mentioned above can aid in establishing a threshold for identifying the source of a text. Furthermore, the categorization provided by the professors serves as additional metadata that can extend the tool's accuracy. This stems from the notion that each category may possess a distinct cut-off threshold for the perplexity score for distinguishing between human-written and AI-generated texts.

In summary, utilizing a single perplexity threshold can serve as an effective preliminary measure in examining the origins of textual sources. By categorizing texts and applying multiple perplexity thresholds, we can improve our tool's ability to analyze and differentiate these origins accurately.

\subsection{Main Algorithm of HowkGPT}
In natural language processing, encoding represents text as numerical vectors that can be used as input to a machine learning model. Embeddings are commonly used in NLP to represent words or phrases as dense vectors of real numbers. These vectors are learned by a neural network during the training process based on the relationships between the words or phrases in a corpus of text. Each model has its own maximum length of tokens that it can represent in embeddings, and this is specific to the architecture followed, e.g., 1024 tokens for GPT2 model~\cite{gpt2-model}.

Algorithm~\ref{alg:calculate_ppl} describes the steps followed in order to calculate perplexity. The actual implementation is done in Python due to several frameworks and libraries supporting machine learning-related projects. Line~15, specifically, retrieves the encodings of the tokens for a specific range of the input text. Encodings (Embeddings) are the representations of tokens in numerical vectors that can be used from a machine-learning model. The result is cloned in the following line because the original values are needed in their initial state in each iteration. The clone function gets a deep copy of the values and stores them in a separate memory location. Whatever is done in the cloned instance does not affect the initial variable value. Moving on to line~18, we retrieve the model output given a specific range of the encodings, which includes the cross entropy loss that is the main factor for perplexity calculation. Consequently, in the next lines of code, the algorithm retrieves the loss and appends it to the array (nnls) which holds all the losses. The whole process runs in a loop in order to cover all text provided as input.

\begin{algorithm}[!t]
\caption{Calculate Perplexity}
\label{alg:calculate_ppl}
\begin{algorithmic}[1]
    \Require~~\\
    $STRIDE$: length of processing window\\
    $M\_LEN$: length of maximum processing window, depended to model\\
    $model$: the gpt2 pretrained model\\
    $tokenizer$: the gpt2 pretrained model tokenizer\\
    $text$: the text to analyze
    \Statex
    \Function{calculate\_perplexity}{$text$}
        \State $seq\_len, encodings \leftarrow$ tokenizer($text$) \Comment{Get array of text representations in the model space and sequence length using text as input.}
            
        \State $nlls \leftarrow []$ \Comment{negative log likelihoods - empty list}
        \State $prev\_end\_loc \leftarrow 0$
        \State $end\_loc \leftarrow 0$
        \State $begin\_loc \leftarrow 0$

        \While{$begin\_loc \leq seq\_len$}
            \State $end\_loc \gets \min(begin\_loc + M\_LEN, seq\_len)$
            \State $trg\_len \gets end\_loc - prev\_end\_loc$
            \State $input\_ids \gets encodings[begin\_loc : end\_loc]$
            \Comment{Retrieve text representation in model space for range}
            \State $target\_ids \gets clone(input\_ids)$
            \State $target\_ids[0 : array\ length\ -\ trg\_len] \leftarrow -100$
            \Comment{Exclude the specific values from the following calculations, thus mark them with -100}
            \State $outputs \gets model(target\_ids$)\Comment{Get model output for each token in selected range}
            \State $nlls.append(outputs.loss)$
            \If{$end\_loc$ = $seq\_len$}
                \State \textbf{break}
            \EndIf
            \State $prev\_end\_loc \gets end\_loc$
            \State $begin\_loc \gets begin\_loc + M\_LEN$
        \EndWhile

        \Return {$e^{mean(nlls)}$}
    \EndFunction
\end{algorithmic}
\end{algorithm}

In order to better understand the functionality of the proposed algorithm we provide detailed step by step intermediate results for the use cases mentioned in Table~\ref{tab:dataset_flavors}. Table~\ref{tab:moving_window} depicts ChatGPT-generated and student written text state and parameter values in the intermediate operations of Algorithm~\ref{alg:calculate_ppl}. Column \textit{begin\_loc} depicts the start location of the moving window, column \textit{end\_loc} is the end of it and \textit{trg\_len} is the actual window length, which is not always the same because for example we might find the end of the available text. The parameters are the ones that essentially define how the moving window is being selected in each iteration, Table~\ref{tab:moving_window} additionally depicts the actual text that corresponds to the calculated values of these parameters. Regarding student replies the iterations are clipped because the text is much longer and the table's objective is to depict the method and not expand all iterations. Before moving on, it is important to note that when a model tries to generate a sentence it uses the same logic as we do here in order to calculate perplexity. The calculation of perplexity is like inspecting what happened, what was the log likelihood, when the tokens were generated.

In order to provide a better analysis, we changed the default GPT2 max window length from one thousand twenty four down to sixty four and stride to thirty two. This was done in order to force the algorithm to run some iterations on the texts we have, which in their majority are less than one thousand twenty four tokens. Text is the portion of the answer that is used as context from the model. The grayed out text marks the overlapping parts in each iteration. This happens since the model needs context to predict tokens (in our case calculate log likelihood), and the more it has before the token is generated, the better. Small step size induces more overlapping, which leads to having bigger context that eventually contributes in the calculation of the log likelihood of the next token. The ideal step would be just one token, but this means reduced performance because we need more iterations. If step is same as the max window size it means there will be no common context on deciding the next token. In a sentence or paragraph, the text is interconnected, each token is selected-written based on what tokens exist before. Eventually, if we decide not to take into consideration the previous tokens when trying to see the log likelihood of the next one means, we ignore this interconnection. Thus, the overlapping behavior is desired and the amount of overlapping is up to the task and performance we want our model to have. The \textit{nll} parameter is the negative log likelihood value retrieved on each iteration, excluding the overlapping part in order not to add the loss twice. This is eventually appended to the \textit{nnls} array. Finally, the exponentiated average of \textit{nnls} results in the perplexity of the whole text. Conclusively, it is clear that perplexity of student written text is significantly higher to its total, but also to each context used in each iteration.

\sethlcolor{lightgray}

\begin{table*}[!t]
\caption{Parameter values for each iteration of Algorithm~\ref{alg:calculate_ppl}, grayed out text marks the overlapping context with the previous iteration of the moving window.}
\label{tab:moving_window}
\begin{tabularx}{\textwidth}{| L | c | c | c | c |}
\hline
\multicolumn{1}{| c |}{\textbf{contextual text}} & \textbf{begin\_loc} & \textbf{end\_loc} & \textbf{trg\_len} & \textbf{nll} \\
\hline
\multicolumn{5}{| c |}{\textbf{ChatGPT Reply}} \\
\hline
The cyber-threat landscape is constantly evolving, but some common types of threats include malware, phishing, ransomware, and distributed denial of service (DDoS) attacks. Malware refers to malicious software that can infect a computer or device and allow an attacker to gain access to sensitive information or disrupt operations. Phishing refers &
  0 & 64 &  64 & 2.338 \\
\hline
\hl{attacks. Malware refers to malicious software that can infect a computer or device and allow an attacker to gain access to sensitive information or disrupt operations. Phishing refers} to attempts to trick individuals into providing sensitive information, such as passwords or credit card numbers, through fraudulent email or website. Ransomware is a type of malware &
  32 & 96 & 32 & 1.938 \\
\hline
\hl{to attempts to trick individuals into providing sensitive information, such as passwords or credit card numbers, through fraudulent email or website. Ransomware is a type of malware} that encrypts a victims files and demands payment in exchange for the decryption key. DDoS attacks involve overwhelming a website or network with traffic to disrupt service &
  64 & 128 & 32 & 2.774 \\
\hline
\hl{that encrypts a victims files and demands payment in exchange for the decryption key. DDoS attacks involve overwhelming a website or network with traffic to disrupt service}. The motivation behind using cyber-attacks instead of physical attacks is often the ability to carry out an attack remotely and with a lower risk of being caught. Cyber &
  96 & 160 & 32 & 2.904 \\
\hline
\hl{. The motivation behind using cyber-attacks instead of physical attacks is often the ability to carry out an attack remotely and with a lower risk of being caught. Cyber}-attacks can also be more cost-effective and have a greater potential impact than physical attacks. Additionally, many organizations and individuals have valuable information stored electronically, making &
  128 & 192 & 32 & 2.399 \\
\hline
\hl{-attacks can also be more cost-effective and have a greater potential impact than physical attacks. Additionally, many organizations and individuals have valuable information stored electronically, making} it a more attractive target for cybercriminals. Additionally, it is also a way for hackers to disrupt services of a company or government without having a physical presence &
  160 & 224 & 32 & 2.6 \\
\hline
\hl{it a more attractive target for cybercriminals. Additionally, it is also a way for hackers to disrupt services of a company or government without having a physical presence} in the location. &
  192 & 228 & 4 & 1.898 \\
\hline
\multicolumn{5}{| c |}{\textbf{Student Reply}} \\
\hline
With the rise of the IoT and crypto industries, the development of AI and machine learning, and frankly, the inescapable digitalization of every aspect of our lives, the digital world finds itself in a rather troubling situation. There are various cyber threats that pose threat to computer systems nowadays and it does not seem that their &
0 & 64 & 64 & 3.201 \\
\hline
\hl{lives, the digital world finds itself in a rather troubling situation. There are various cyber threats that pose threat to computer systems nowadays and it does not seem that their} number is going to diminish. Cyber worms, botnets, rootkits and backdoors, different types of ransomwares, DDoS attacks, spam, &
32 & 96 & 32 & 3.14 \\
\hline
\hl{number is going to diminish. Cyber worms, botnets, rootkits and backdoors, different types of ransomwares, DDoS attacks, spam,} and phishing (especially email phishing), trojans, backdoors, etc. – all the after-mentioned vulnerabilities pose a tremendous threat to computer systems &
64 & 128 & 32 & 3.313 \\
\hline
.. & .. & .. & .. & .. \\
\hline
\hl{critical infrastructure is successful it might as well attract as much attention as physical attacks. Scalability is another parameter that plays a vital role in any cyber-attack.} To scale up a physical attack, would require bringing additional troops, and military equipment, which can be extremely troublesome in logistical terms, especially during a battle. Whereas &
704 & 768 & 32 & 4.256 \\
\hline
\hl{To scale up a physical attack, would require bringing additional troops, and military equipment, which can be extremely troublesome in logistical terms, especially during a battle. Whereas} in the case of cyber-attacks, one might argue that scalability is relatively simpler, as it implies the deployment of more computing resources, which can be done &
736 & 800 & 32 & 3.18 \\
\hline
\hl{in the case of cyber-attacks, one might argue that scalability is relatively simpler, as it implies the deployment of more computing resources, which can be done} quicker than say shipment of additional equipment to the battlefield. This, however, does not imply it is easy in all terms. &
768 & 825 & 25 & 3.799 \\
\hline

\end{tabularx}
\end{table*}

\subsection{Application Overview}
\begin{figure}[!t]
\centering
\subfloat[Offline Process]
{\includegraphics[width=\columnwidth]{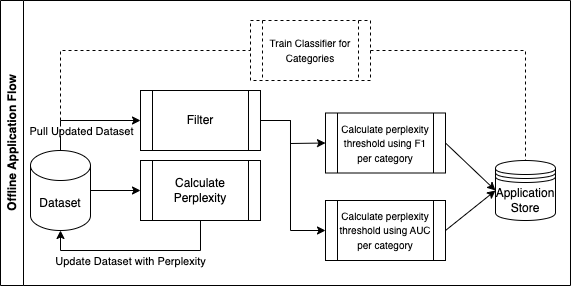}\label{fig:offline_process_flow}}
\hfill
\begin{center}
\subfloat[Live Process]
{\includegraphics[width=0.8\columnwidth]{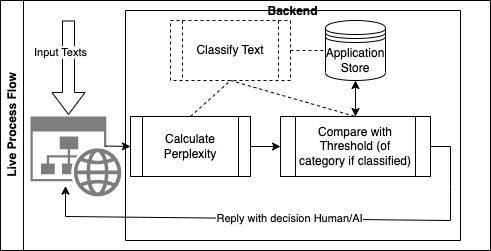}\label{fig:live_process_flow}}
\end{center}
\caption{Offline and Live process flows of the application.}
\end{figure}

The first part of the application consists of an offline process as shown in Figure~\ref{fig:offline_process_flow}. Initially for each text in the dataset, perplexity is calculated and stored back to the dataset. When the whole dataset is updated, two parallel processes calculate the perplexity thresholds with or without using the categorization provided by the taxonomy for each method F1 and AUC. In order to do the evaluation following in Section~\ref{sec:evaluation} the \textit{Filter} process is used. Filter is called multiple times in order to produce the different dataset flavors (Table~\ref{tab:dataset_flavors}). Eventually all outputs are stored in the applications storage. The storage can then be used by the live process flow.

The dashed line represents the classification process employed by HowkGPT to identify the categorization of an input text. The classification process is a neural network that uses the dataset mentioned in Section~\ref{sec:dataset_construction} and classifies the text into categories according to the taxonomy mentioned above. However, the classification function is not activated as a default setting in the application due to its performance being constrained by the limited size of the dataset. The classifier requires further optimization and fine-tuning on a more enriched dataset to enhance its effectiveness.

Secondly, the live process consists of a web application and a backend microservice. The web application accepts texts through a web form and forwards them to the backend service that calculates the perplexity in real time, compares the value with the threshold and replies with the origin of the text, human-written or AI-generated. The \textit{Calculate Perplexity} process is the same used in the offline process. Again the dashed entities are not in the pipeline by default. In case they are enabled the neural network classifier would provide the category of the text and then the comparison would be performed on the adjusted perplexity threshold of the category.

%% file: sections/5_evaluation.tex
\subsection{Dataset Exploration}\label{sub:dataset_flavors}
As mentioned before, the dataset comprises questions about mathematical concepts, coding, and specific responses that are notably brief. Considering the fundamental principles of the perplexity metric, such short texts do not provide adequate context for the algorithm to provide a meaningful value. Similarly, mathematics and coding responses employ common patterns, making it highly likely that the responses to this kind of question would bear strong similarities no matter who wrote or generated the response. A series of experiments involving data filtering was conducted to conclude that it is prominent to differentiate texts containing code and/or mathematical content using the perplexity metric, given the characteristics of the present dataset. This, however, remains a topic of discussion that warrants further investigation and validation. The provided taxonomy offered a straightforward approach to filter the dataset in line with the aforementioned reasoning, thereby facilitating the exploration of various techniques toward our ultimate objective of establishing the perplexity threshold. Further investigations could potentially lead to more flavors and filtering based on the actual content, such as excluding special characters or bulleted text, which may have peculiar effects on the perplexity. This is also an aspect that may be explored in future work. Conclusively, the dataset flavors is a first attempt to exclude texts that induce noise and anomaly behavior in the perplexity calculation. The different flavors of the dataset are listed in Table~\ref{tab:dataset_flavors} along with notations which will be used throughout the paper to identify the flavor. 

\begin{table}[!t]
\caption{Dataset Flavors for Filtering}
\label{tab:dataset_flavors}
\centering
\begin{tabular}{ | c | c | }
\hline
\textbf{Flavor} & \textbf{Notation}\\
\hline
Original & orig\\
\hline
With text more than 250 characters long & $\geq 250$\\
\hline
Without math related questions & !math\\
\hline
Without code related questions & !code\\
\hline
Without code and math related questions & !math !code\\
\hline
\end{tabular}
\end{table}

\begin{figure}[!t]
\centering
\subfloat[Original dataset]
{\includegraphics[width=0.8\columnwidth]{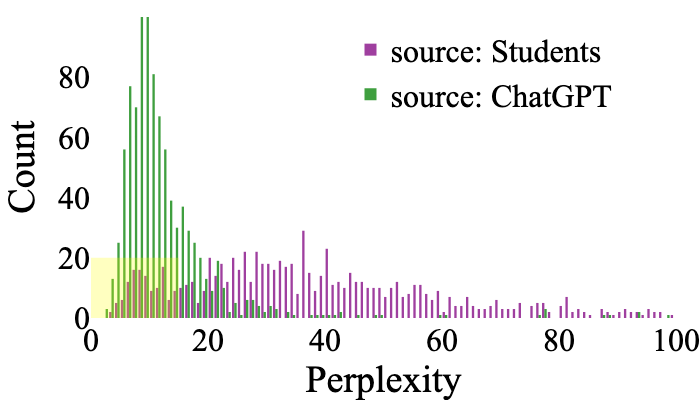}\label{fig:ppl_student_gpt_hist_original}}
\hfill
\subfloat[$\geq 250$]
{\includegraphics[width=0.49\columnwidth]{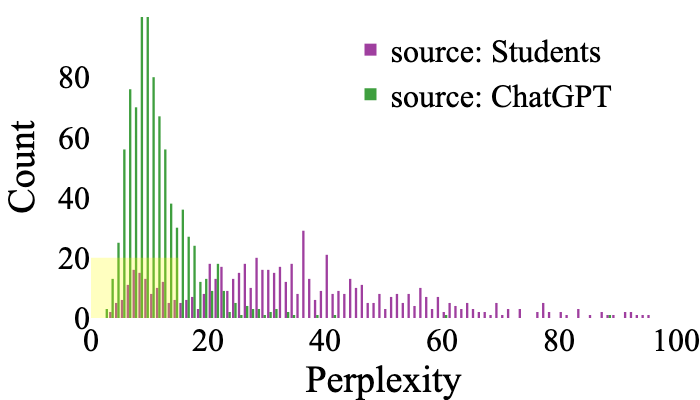}\label{fig:ppl_student_gpt_hist_lt_250}}
\hfill
\subfloat[!code]
{\includegraphics[width=0.49\columnwidth]{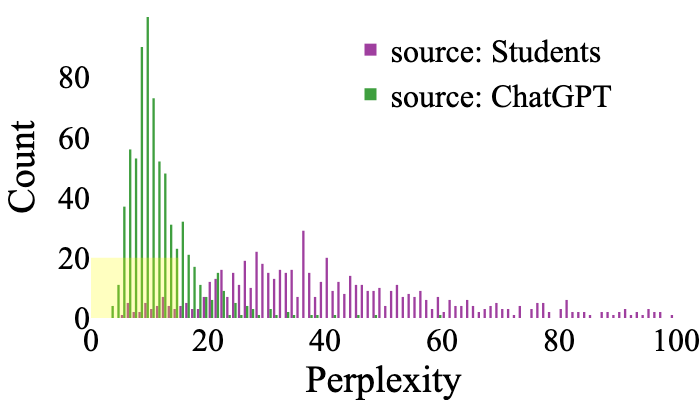}\label{fig:ppl_student_gpt_hist_no_code}}
\hfill
\subfloat[!math]
{\includegraphics[width=0.49\columnwidth]{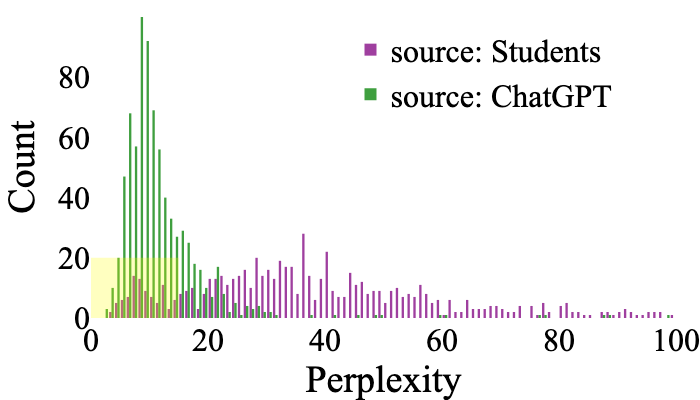}\label{fig:ppl_student_gpt_hist_no_math}}
\hfill
\subfloat[!math !code]
{\includegraphics[width=0.49\columnwidth]{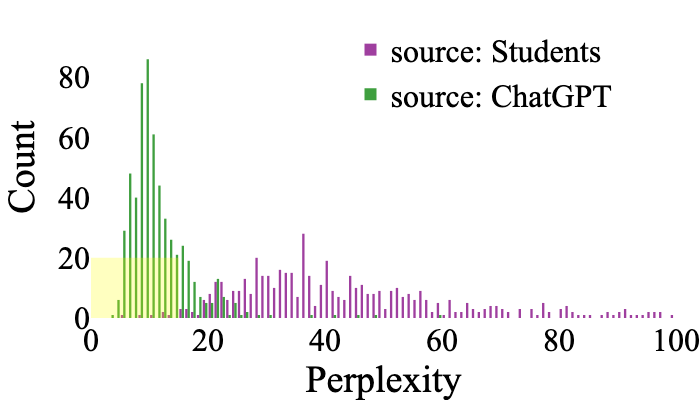}\label{fig:ppl_student_gpt_hist_no_math_no_code}}
\hfill
\caption{Distributions of perplexity values for the original dataset and also after applying different filtering strategies as mentioned in Table~\ref{tab:dataset_flavors}.}
\label{fig:ppl_student_gpt_hist}
\end{figure}

In order to have a better understanding, Figure~\ref{fig:ppl_student_gpt_hist} depicts exactly how the perplexity values for the responses are distributed for each class (i.e., Students and ChatGPT), and highlights with yellow color an area of interest for each flavor. Each sub-figure is generated using the corresponding dataset flavor mentioned in Table~\ref{tab:dataset_flavors}. The yellow area is the part of the distribution that is mainly affected by the filtering happening in each dataset flavor and thus highlighted. It spreads from zero up to twenty value of perplexity, which is conveyed by inspection of the distributions. Original distribution in Figure~\ref{fig:ppl_student_gpt_hist_original} shows that the majority of ChatGPT generated answers have low perplexity while student written replies are distributed more evenly and the majority seems to be distributed above fifteen to twenty perplexity value. The way that the responses are distributed will affect what will be the value of the perplexity threshold which will define the classification of text to human written or AI-generated. Distribution~\ref{fig:ppl_student_gpt_hist_lt_250} has a small decrease in student written instances around fifteen perplexity, which implies that there are not a lot of texts with low perplexity and less than two hundred fifty characters length. The rest of the distribution seems intact which also applies for the AI-generated replies. On the other hand, Figure~\ref{fig:ppl_student_gpt_hist_no_code} and Figure~\ref{fig:ppl_student_gpt_hist_no_math} display a huge impact in the area highlighted with yellow and this is explained by the common pattern usage that happens in mathematical and coding questions. This impact is translated as a small decrease for AI-generated and a bigger one for student written replies. Inevitably, the last flavor depicted in Figure~\ref{fig:ppl_student_gpt_hist_no_math_no_code} has the lowest instances of student written solutions in the highlighted area. It is important here to note that some answers have perplexity scores way above one hundred but were excluded for clarity.

\subsection{Statistical Exploration of Threshold}\label{sub:explore_threshold}

In this section we provide a mathematical/statistical verification in order to investigate the optimal threshold value of perplexity to identify the origin of any responses. In our case, since we have to make a decision based on a metric with two different groups or classes (text source: human or AI) we can leverage the receiver operating characteristic curve (ROC curve) and the area under curve (AUC). ROC curve and AUC methods are mostly used when given a classifier, we have a set of possibilities for each class. This means that the curve will have multiple points on the (x, y) plain for each threshold value~\cite{BROWN200624}. In this scenario, there are no probabilities but given a threshold, there is a certainty that source is one of the two classes, this produces only one point on the (x, y) plain along with the root of the axis and (1, 1) point. A high AUC indicates that the method is better at classifying human written samples rather than AI-generated ones, but it does not provide information about the balance between precision and recall. In order to address this behavior, F1 score, which is a measure of a model's accuracy that considers both precision and recall and provides a balanced evaluation of the model's performance, is also considered to be used as a metric for the calculation of the optimal perplexity threshold. The F1 score can be high if the model has high precision and high recall, or if it has a good balance between the two. A high F1 score indicates that the model is performing well in terms of both identifying human written samples and avoiding false positives (FP) and false negatives (FN).

\begin{figure}[!t]
\centering
\subfloat[orig]
{\includegraphics[width=0.6\columnwidth]{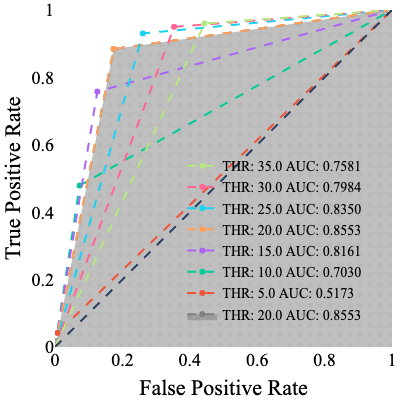}\label{fig:roc_curve_original}}
\hfill
\subfloat[$\geq$ 250]
{\includegraphics[width=0.49\columnwidth]{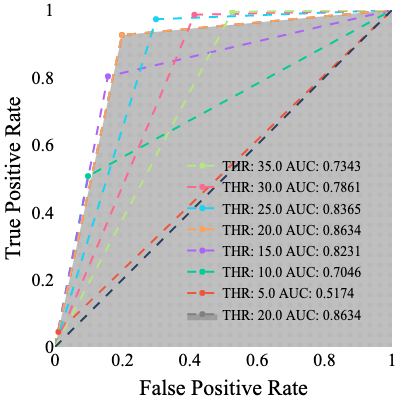}\label{fig:roc_curve_lt_250}}
\vspace{0.2cm}
\subfloat[!code]
{\includegraphics[width=0.49\columnwidth]{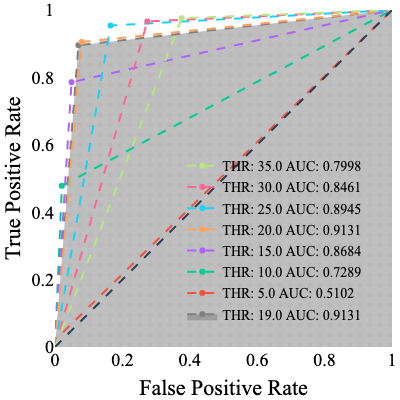}\label{fig:roc_curve_no_code}}
\hfill
\subfloat[!math]
{\includegraphics[width=0.49\columnwidth]{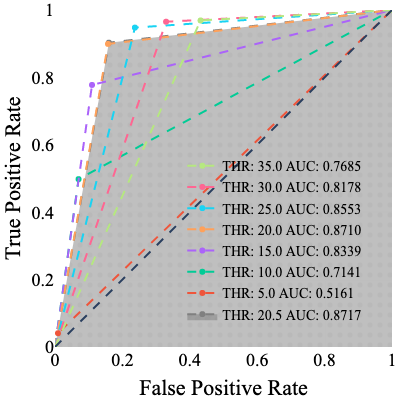}\label{fig:roc_curve_no_math}}
\hfill
\subfloat[!math !code]
{\includegraphics[width=0.49\columnwidth]{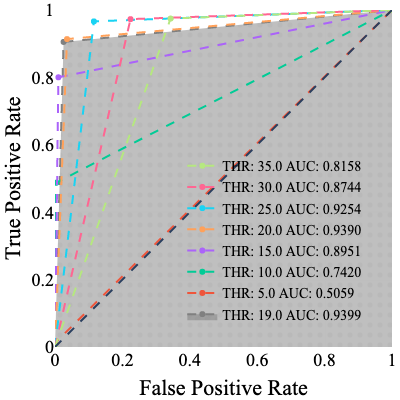}\label{fig:roc_curve_no_math_no_code}}
\caption{ROC curves for different perplexity values (THR). Grayed out regions are the AUC for the optimal threshold. Each sub figure is generated using a dataset flavors.}
\label{fig:roc_curve_auc}
\end{figure}

\begin{figure}[!t]
\centering
\subfloat[orig]
{\includegraphics[width=\columnwidth]{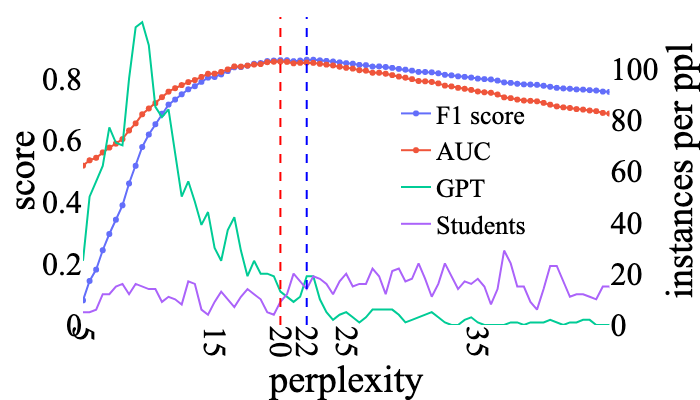}\label{fig:f1_auc_inc_rate_original}}
\hfill
\subfloat[$\geq$ 250]
{\includegraphics[width=0.49\columnwidth]{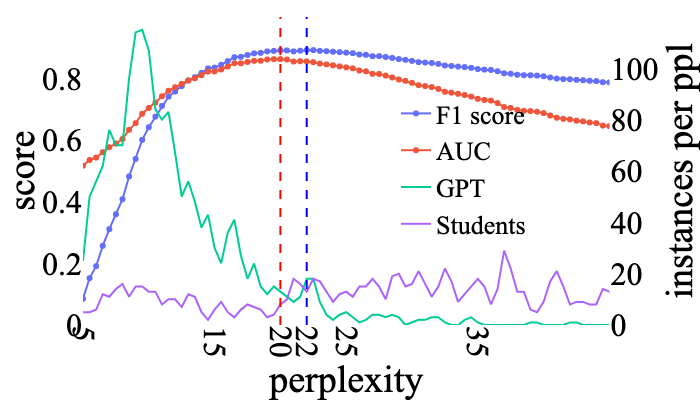}\label{fig:f1_auc_inc_rate_lt_250}}
\vspace{0.2cm}
\subfloat[!code]
{\includegraphics[width=0.49\columnwidth]{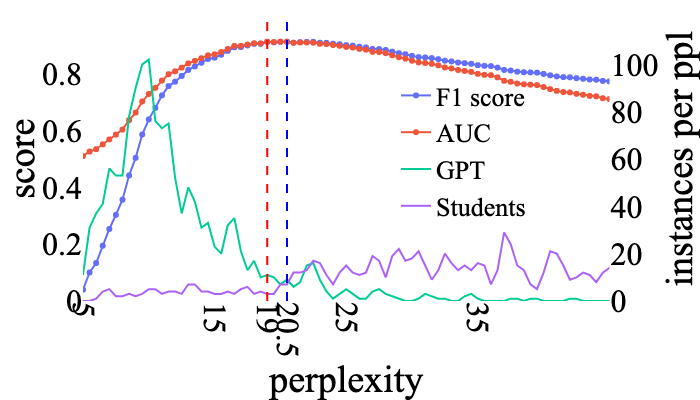}\label{fig:f1_auc_inc_rate_no_code}}
\hfill
\subfloat[!math]
{\includegraphics[width=0.49\columnwidth]{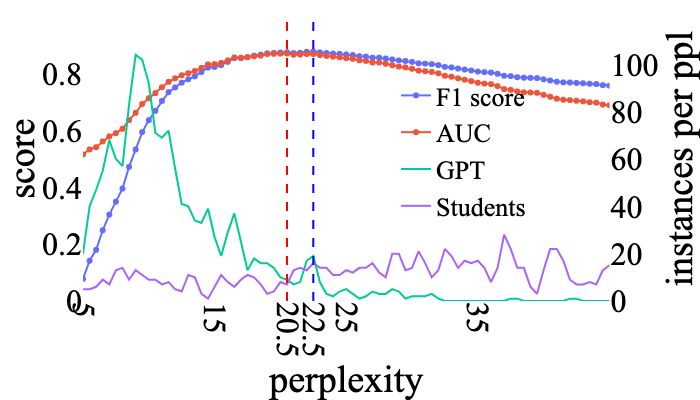}\label{fig:f1_auc_inc_rate_no_math}}
\hfill
\subfloat[!math !code]
{\includegraphics[width=0.49\columnwidth]{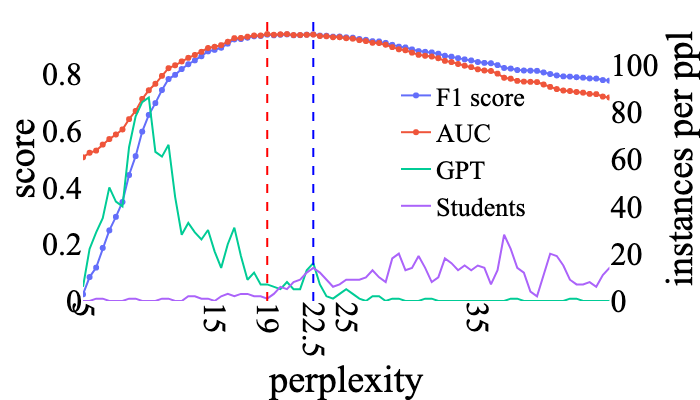}\label{fig:f1_auc_inc_rate_no_math_no_code}}
\caption{F1(blue) and AUC(red) scores, ChatGPT(green) and Students(purple) lines indicate the instances of answers(texts) from each category for a given perplexity. Dashed lines indicate the max F1 and AUC scores and the corresponding perplexity values. Each sub figure is generated using a dataset flavors.}
\label{fig:f1_auc_inc_rate}
\end{figure}

ROC curves for the original data and all flavors are depicted in Figure~\ref{fig:roc_curve_auc}. For each perplexity value we have a different curve, a few of them are manually picked and displayed here. The dominant area (AUC), bigger area, is grayed out in the figures and identifies the optimal perplexity threshold for classifying texts to either human written or AI-generated. The variations due to the different dataset flavors can be identified by the (x,y) position of the unique point that essentially defines the ROC curves in Figure~\ref{fig:roc_curve_auc}. The best results in our case can also be retrieved if we consider the distance to (0,1) point in the (x,y) plane. Figure~\ref{fig:roc_curve_no_math_no_code} has the biggest AUC. In order to have a better picture of these results, we visualize in Figure~\ref{fig:f1_auc_inc_rate_original}, Figure~\ref{fig:f1_auc_inc_rate_lt_250}, Figure~\ref{fig:f1_auc_inc_rate_no_code}, Figure~\ref{fig:f1_auc_inc_rate_no_math} and Figure~\ref{fig:f1_auc_inc_rate_no_math_no_code} the AUC values along with F1 scores. In parallel, with purple and green are displayed the instances of answers that have the given perplexity value, for students and ChatGPT generated text accordingly. Essentially, we combine in one figure AUC values, F1 scores, the optimal values for each metric and finally the histograms in Figure~\ref{fig:ppl_student_gpt_hist} visualized in a more continuous way. Along with this we have the F1 score in blue and AUC value in red for each perplexity value (x axis). The two metrics (AUC and F1 scores) give back different perplexity suggestions for the same dataset flavor. The deviation can be explained if we consider what these two scores represent as explained in previous paragraph.

The way that these metrics are calculated can explain why they differentiate in the choice of optimal perplexity threshold. Taking as example the case of Figure~\ref{fig:f1_auc_inc_rate_no_math_no_code} the F1 score indicates a perplexity threshold value of 22.5 while AUC points to 19. As we discussed previously, F1 score tends to avoid FPs and FNs which explains the highest threshold compared to AUC which is a more balanced metric between true positive rate (TPR) and false positive rate (FPR). In !math !code dataset flavor, student samples in low perplexities increase with a very slow rate compared to the other plots and thus gives the opportunity to F1 score to maximize its value later, in a bigger perplexity value compared to the other dataset flavors.

Considering the limited size dataset, it can be claimed that F1 and AUC methods give different results for the perplexity threshold and the distance between them can be even broader in case of a different dataset. We will further explore this in future study with a more enriched dataset. In the next subsection, we will analyze how these two methods adapt in different classified texts given the categorizations in Table~\ref{tab:dimension_subcategory_explanation}.

\subsection{Threshold per Category}
The categorization of the dataset creates an opportunity to explore if the grouping can give better classification accuracy while finding the optimal threshold using the AUC and F1 score methods. We split the dataset to train (90\%) and test (10\%), calculate the optimal threshold per category (Table~\ref{tab:thr_per_cognitive_per_dataset} and Table~\ref{tab:thr_per_knowledge_per_dataset}) on the train dataset and then calculate the classification accuracy on the test set. For both dimensions, knowledge in Table~\ref{tab:thr_per_knowledge_per_dataset} and cognitive process in Table~\ref{tab:thr_per_cognitive_per_dataset}, we present all the optimal perplexity thresholds for the different calculation methods (AUC, F1) and for each dataset flavor mentioned in Table~\ref{tab:dataset_flavors}. The optimal perplexity threshold values in many cases are similar and there is no significant spread. There are though, specific cases that the thresholds are significantly bigger than the other cases and this will be explored further in future work.

\begin{table}[!t]
    \centering
    \caption{Threshold per knowledge category per dataset flavor.}
    \begin{tabularx}{\columnwidth}{ | c | c | L | L | L | L | L |}
    \hline
        \multirow{2}{*}{\textbf{Category}} & \multirow{2}{*}{\textbf{Function}} & \multicolumn{5}{c |}{\textbf{Threshold}}\\
         \cline{3-7}
         & & \textbf{orig} & \textbf{$\geq$ 250} & \textbf{!math} & \textbf{!code} & \textbf{!math, !code}\\ 
         \hline
         \multirow{2}{*}{conceptual}    & AUC & 22.0 & 20.5 & 22.5 & 20.5 & 22.5\\ \cline{2-7}
                                        & F1  & 22.5 & 22.0 & 22.5 & 22.5 & 22.5\\
         \hline
         \multirow{2}{*}{factual}       & AUC & 19.0 & 19.0 & 19.0 & 19.0 & 19.0\\ \cline{2-7}
                                        & F1  & 19.0 & 20.0 & 19.0 & 19.0 & 19.0\\
         \hline
         \multirow{2}{*}{procedural}    & AUC & 19.5 & 19.5 & 14.5 & 19.5 & 19.0\\ \cline{2-7}
                                        & F1  & 22.0 & 22.0 & 22.0 & 21.5 & 18.5\\
         \hline
         \multirow{2}{*}{metagognitive} & AUC & 22.5 & 19.0 & 19.0 & 19.0 & 19.0\\ \cline{2-7}
                                        & F1  & 19.0 & 19.0 & 19.0 & 19.0 & 19.0\\
         \hline
    \end{tabularx}
    \label{tab:thr_per_knowledge_per_dataset}
\end{table}

\begin{table}[!t]
    \centering
    \caption{Threshold per cognitive process category per dataset flavor.}
    \begin{tabularx}{\columnwidth}{ | c | c | L | L | L | L | L |}
    \hline
        \multirow{2}{*}{\textbf{Category}} & \multirow{2}{*}{\textbf{Function}} & \multicolumn{5}{c |}{\textbf{Threshold}}\\
         \cline{3-7}
         & & \textbf{orig} & \textbf{$\geq$ 250} & \textbf{!math} & \textbf{!code} & \textbf{!math, !code}\\ 
         \hline
         \multirow{2}{*}{apply}      & AUC & 19.5 & 20.5 & 19.5 & 19.0 & 21.0\\ \cline{2-7}
                                     & F1  & 22.0 & 22.0 & 22.0 & 21.5 & 21.5\\
         \hline
         \multirow{2}{*}{analyze}    & AUC & 17.5 & 17.5 & 27.0 & 17.5 & 27.0\\ \cline{2-7}
                                     & F1  & 25.5 & 17.5 & 27.0 & 25.5 & 27.0\\
         \hline
         \multirow{2}{*}{remember}   & AUC & 20.0 & 22.5 & 20.0 & 20.0 & 20.0\\ \cline{2-7}
                                     & F1  & 22.5 & 22.5 & 22.5 & 22.5 & 22.5\\
         \hline
         \multirow{2}{*}{evaluate}   & AUC & 19.0 & 19.0 & 19.0 & 22.0 & 19.0\\ \cline{2-7}
                                     & F1  & 19.0 & 19.0 & 22.0 & 22.0 & 22.0\\
         \hline
         \multirow{2}{*}{understand} & AUC & 20.5 & 20.0 & 20.5 & 20.5 & 20.5\\ \cline{2-7}
                                     & F1  & 20.5 & 20.5 & 20.5 & 20.5 & 20.5\\
         \hline
         \multirow{2}{*}{create}     & AUC & 15.5 & 15.5 & 23.0 & 25.0 & 25.0\\ \cline{2-7}
                                     & F1  & 31.5 & 25.0 & 31.5 & 25.0 & 25.0\\
         \hline
    \end{tabularx}
    \label{tab:thr_per_cognitive_per_dataset}
\end{table}

\begin{figure}[!t]
\centering
\subfloat[!math !code]
{\includegraphics[width=0.49\columnwidth]{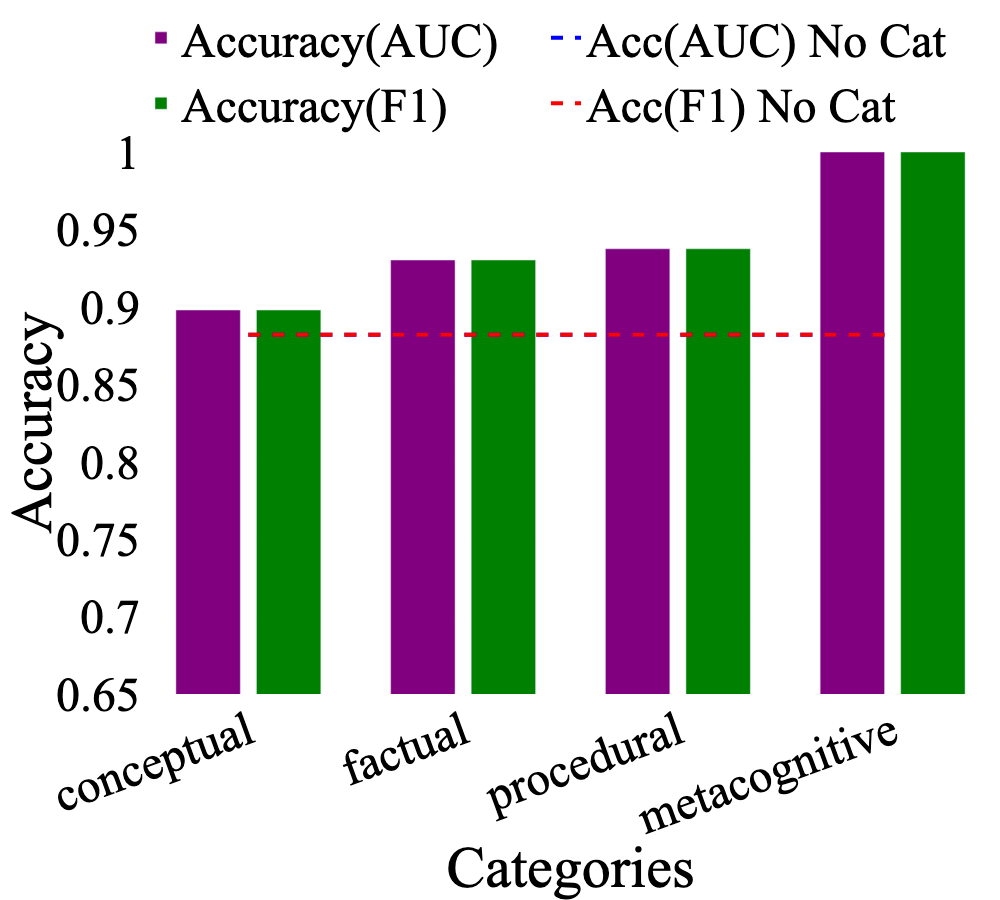}}\label{fig:threshold_accuracy_no_math_no_code_knowledge_dimension}
\hfill
\subfloat[$\geq 250$]
{\includegraphics[width=0.49\columnwidth]{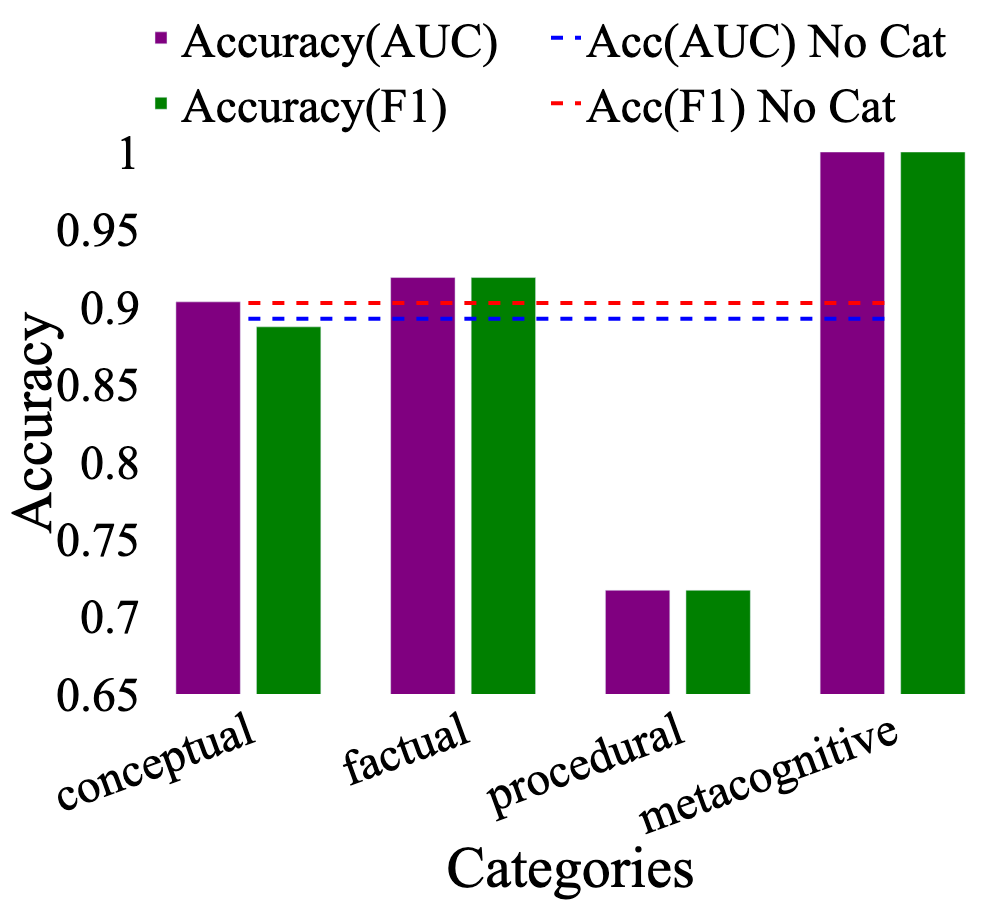}}\label{fig:threshold_accuracy_lt_250_knowledge_dimension}
\hfill
\subfloat[!code]
{\includegraphics[width=0.49\columnwidth]{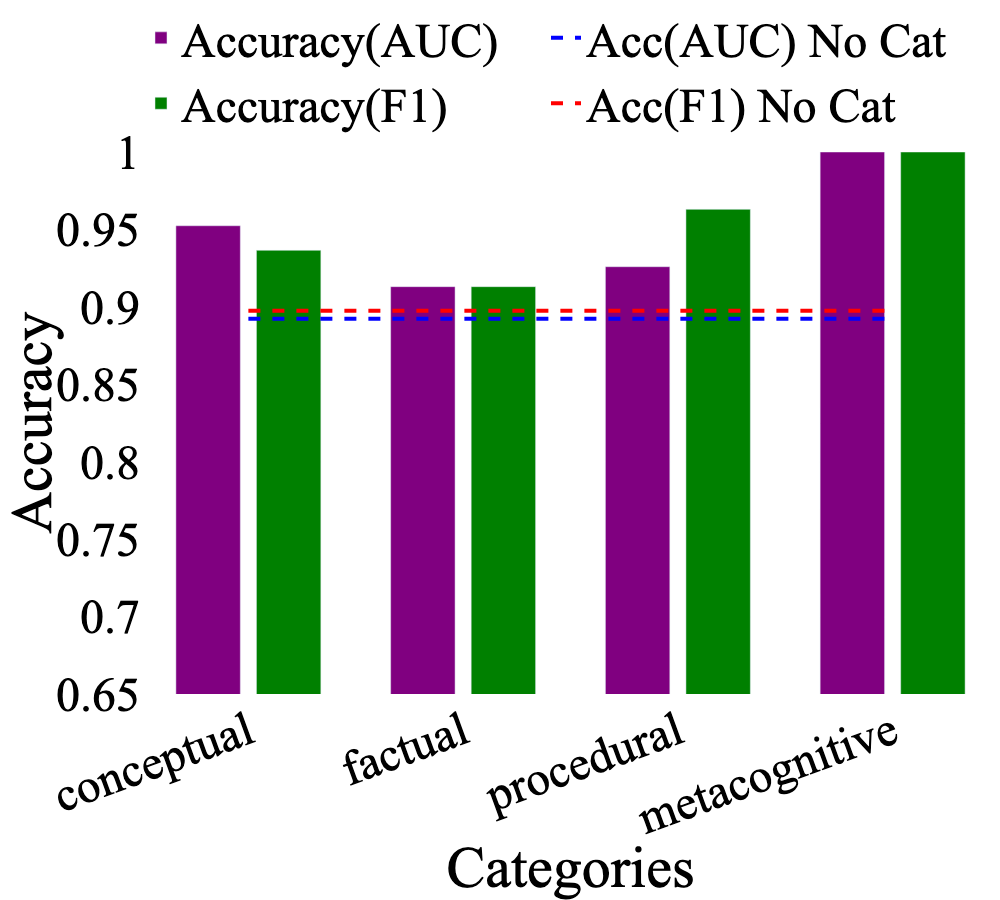}}\label{fig:threshold_accuracy_no_code_knowledge_dimension}
\hfill
\subfloat[!math]
{\includegraphics[width=0.49\columnwidth]{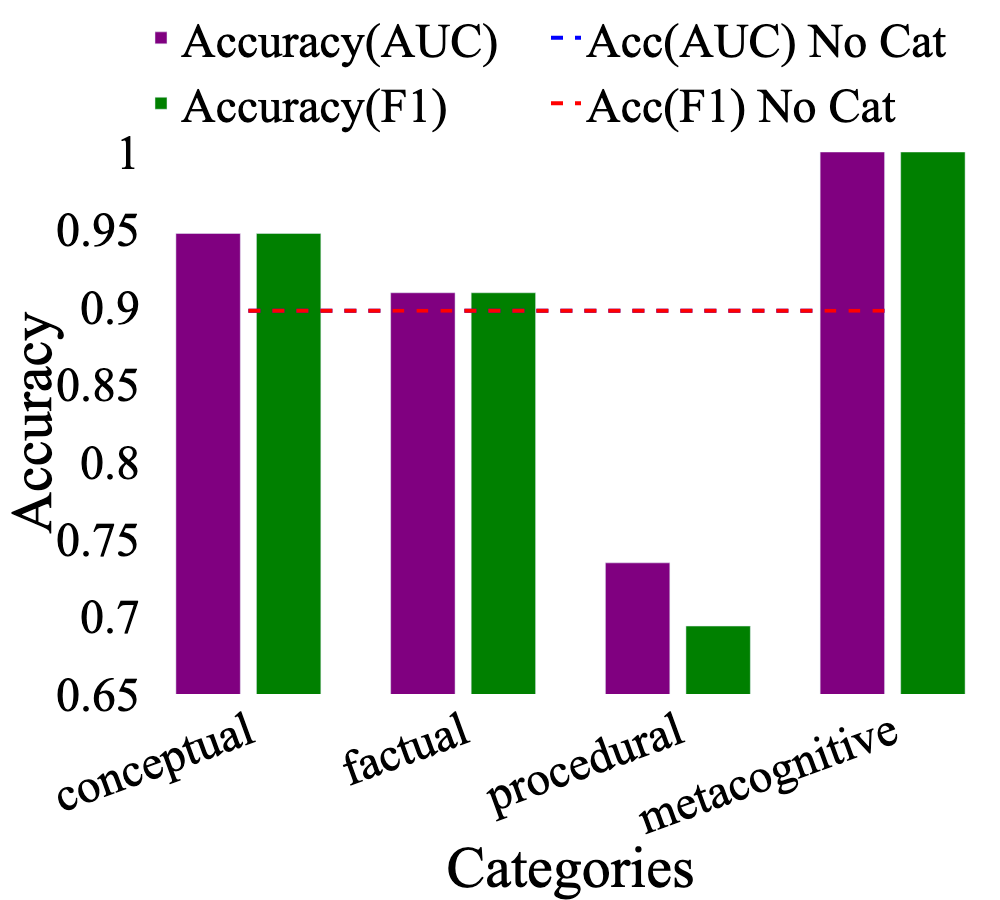}}\label{fig:threshold_accuracy_no_math_knowledge_dimension}
\hfill
\subfloat[orig]
{\includegraphics[width=0.49\columnwidth]{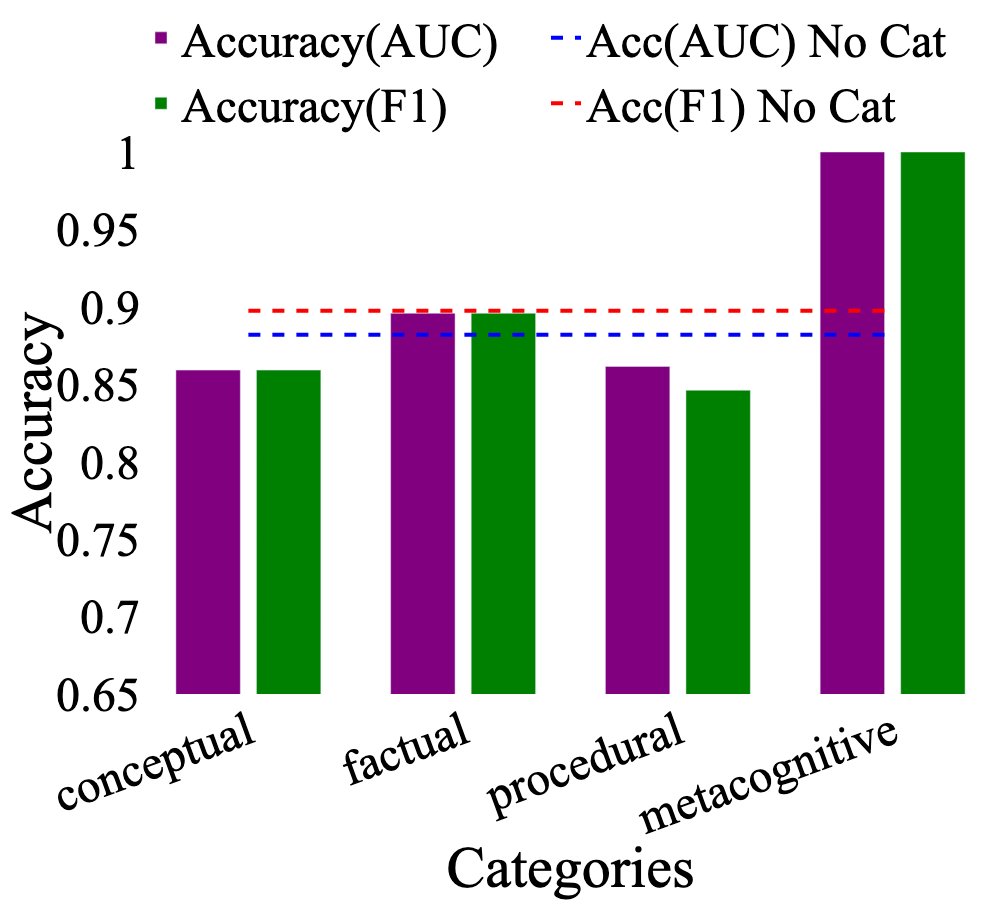}}\label{fig:threshold_accuracy_original_knowledge_dimension}
\hfill
\caption{Accuracy for Knowledge Dimension Subcategories. Purple is accuracy given perplexity threshold calculated by AUC score and green by F1 score. The horizontal red line is the accuracy given AUC score perplexity threshold without using categories and blue given F1 score perplexity threshold.}
\label{fig:accuracy_per_cat_knowledge}
\end{figure}

\begin{figure}[!t]
\centering
\subfloat[!math !code]
{\includegraphics[width=0.49\columnwidth]{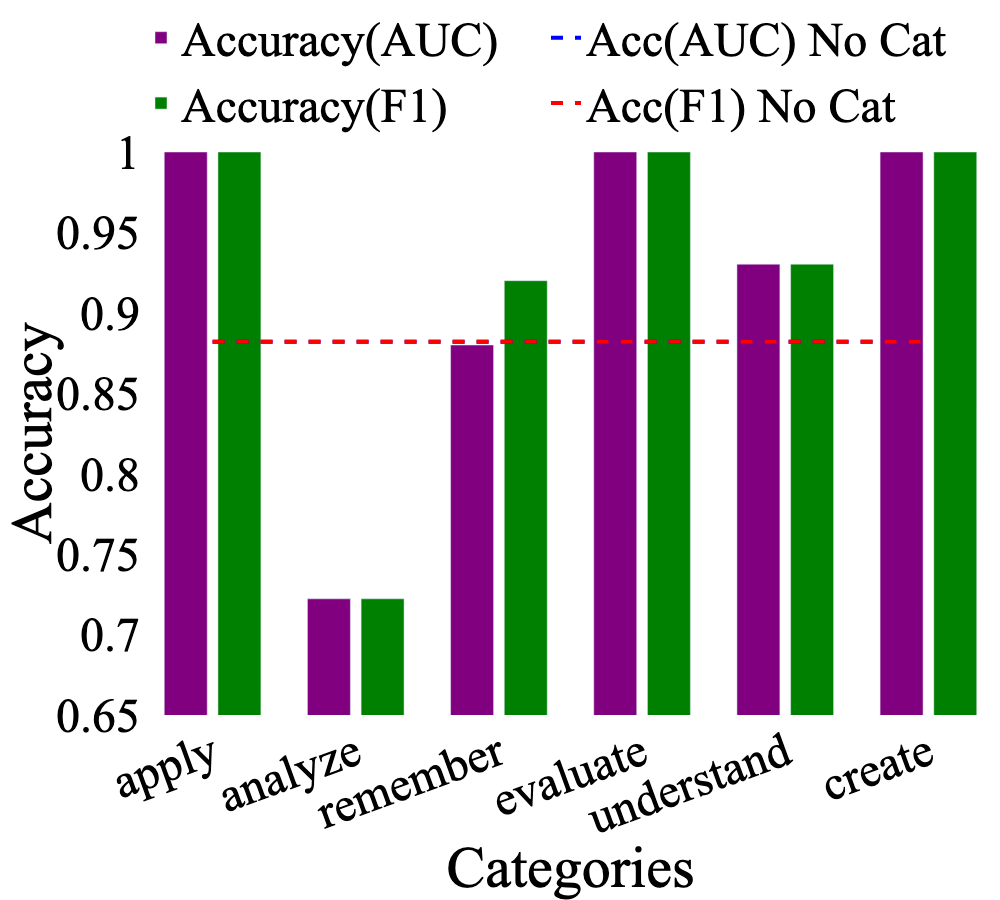}}\label{fig:threshold_accuracy_no_math_no_code_cognitive_process_dimension}
\hfill
\subfloat[$\geq 250$]
{\includegraphics[width=0.49\columnwidth]{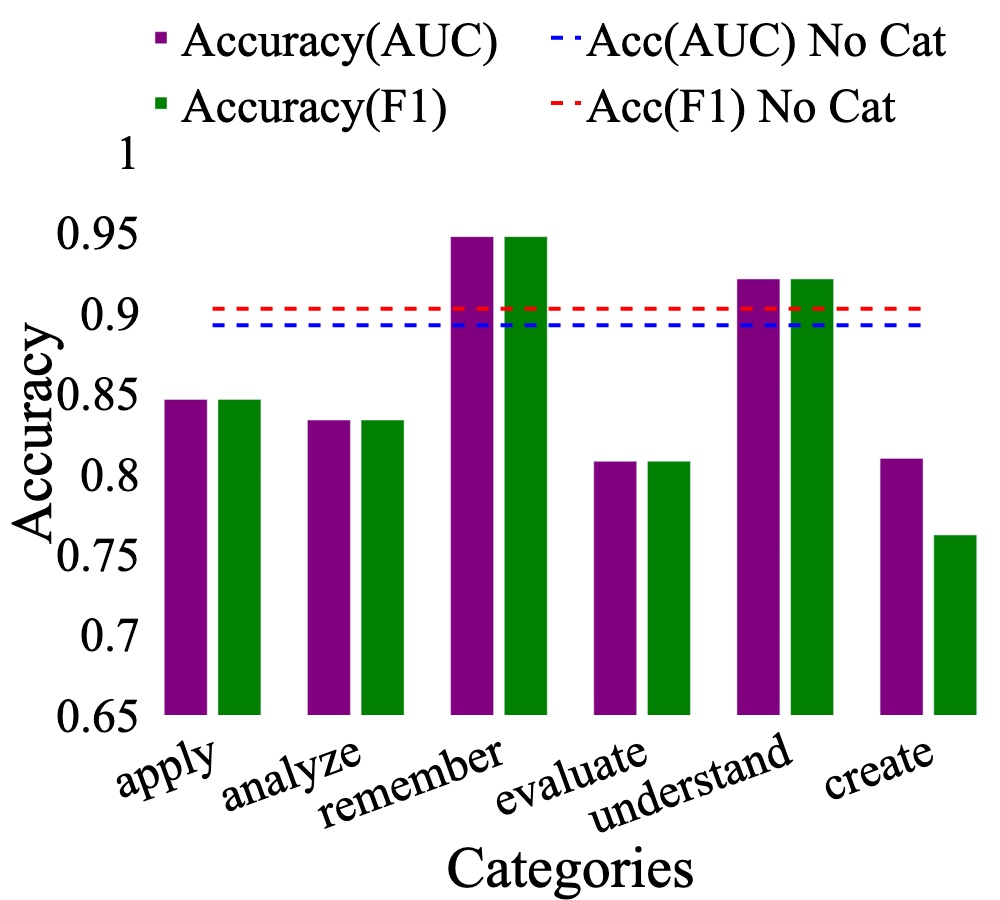}}\label{fig:threshold_accuracy_lt_250_cognitive_process_dimension}
\hfill
\subfloat[!code]
{\includegraphics[width=0.49\columnwidth]{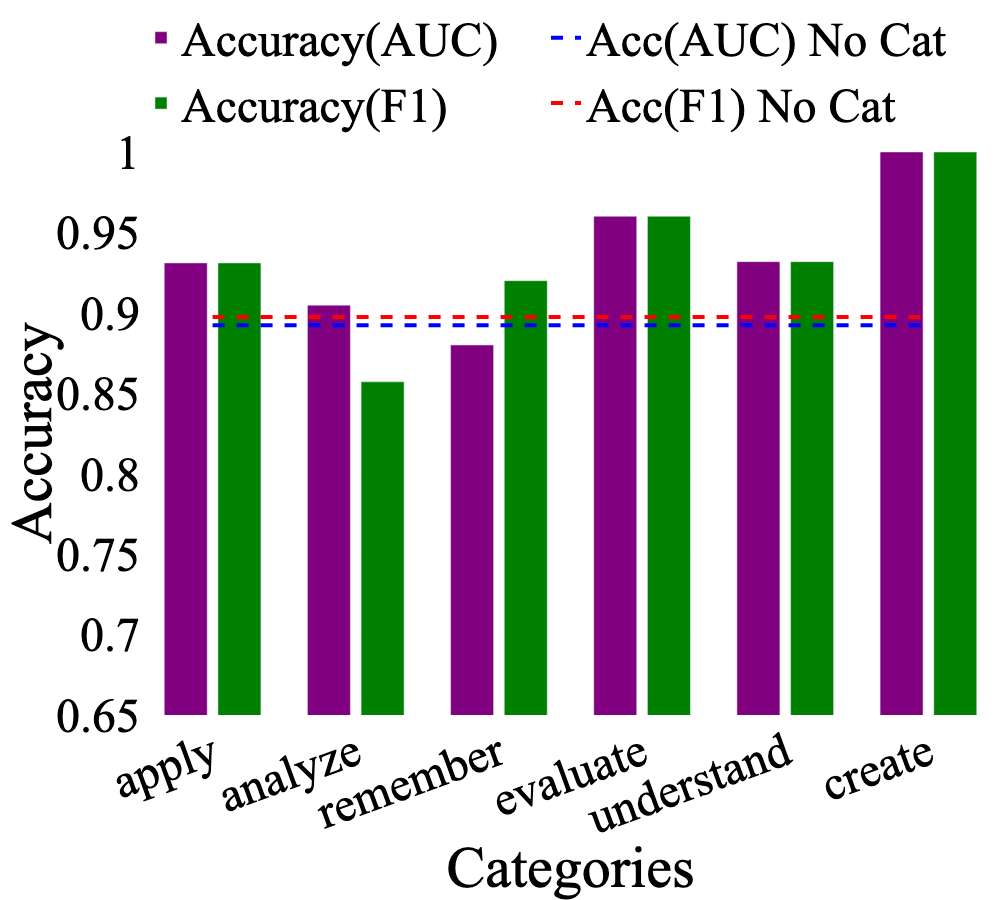}}\label{fig:threshold_accuracy_no_code_cognitive_process_dimension}
\hfill
\subfloat[!math]
{\includegraphics[width=0.49\columnwidth]{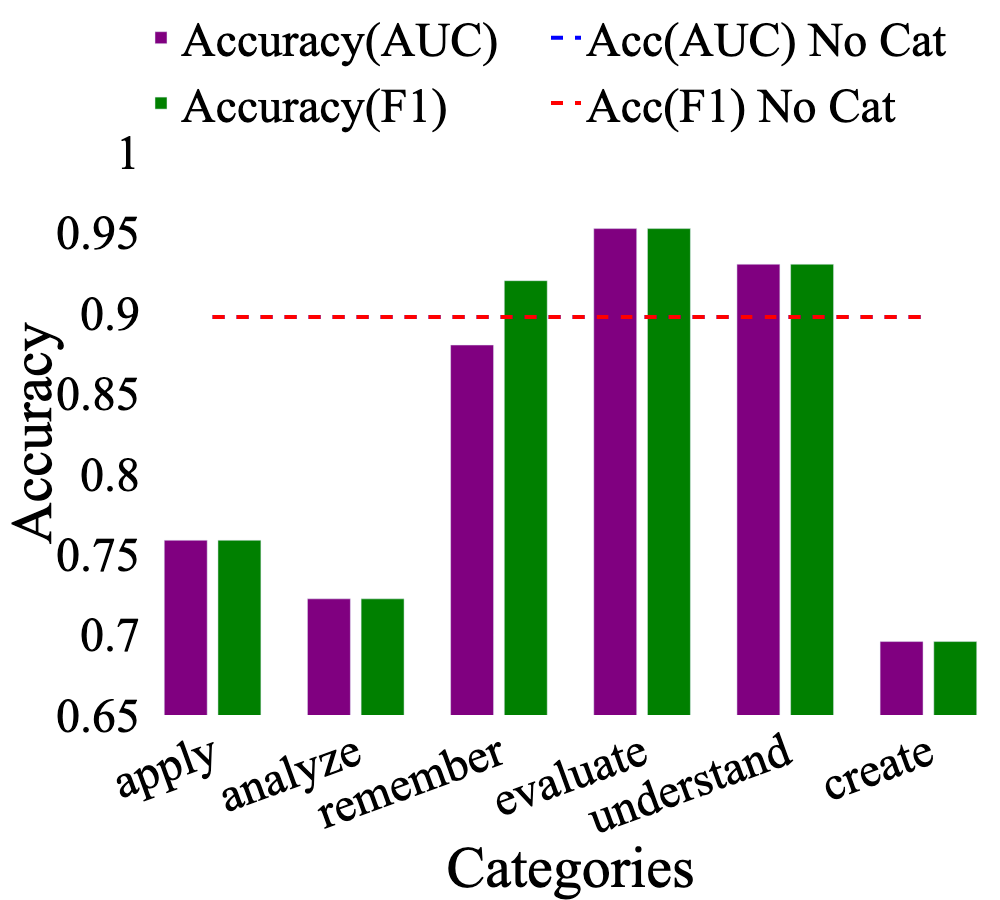}}\label{fig:threshold_accuracy_no_math_cognitive_process_dimension}
\hfill
\subfloat[orig]
{\includegraphics[width=0.49\columnwidth]{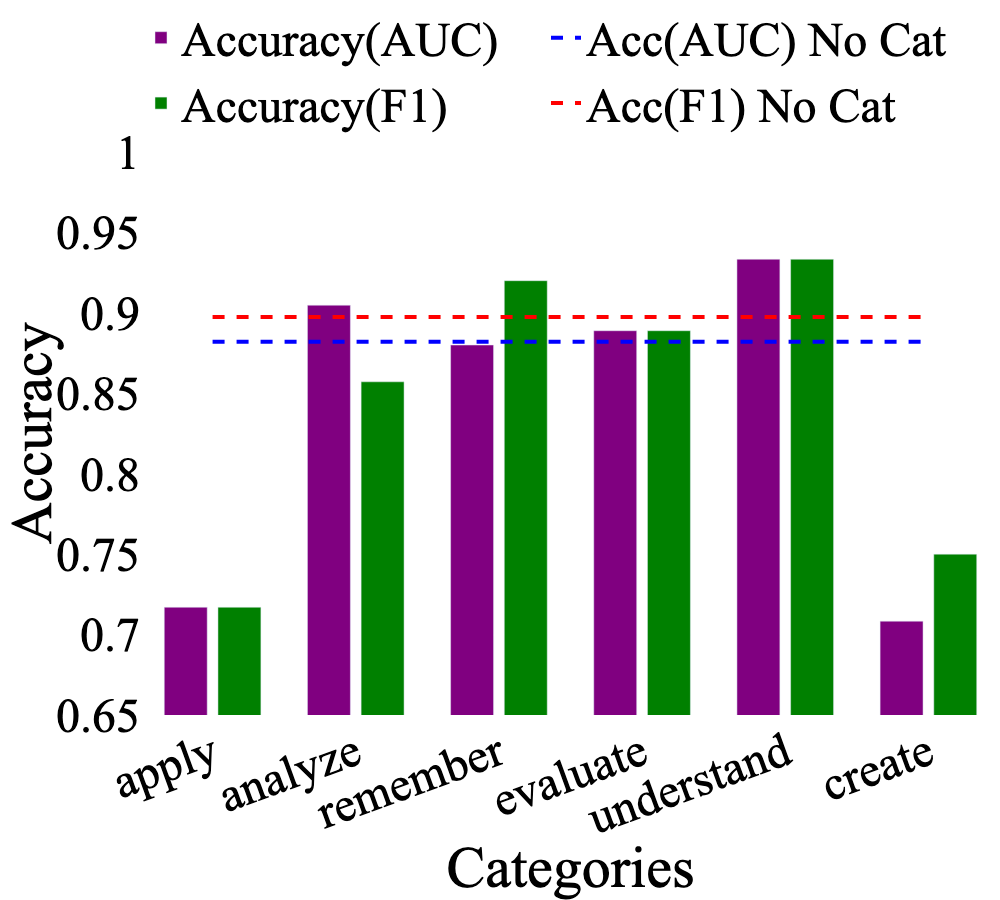}}\label{fig:threshold_accuracy_original_cognitive_process_dimension}
\hfill
\caption{Accuracy for Cognitive Process Dimension. Purple is accuracy given perplexity threshold calculated by AUC score and Green by F1 score. The horizontal Red line is the accuracy given AUC score perplexity threshold without using categories and Blue given F1 score perplexity threshold}
\label{fig:accuracy_per_cat_cognitive}
\end{figure}

Figure~\ref{fig:accuracy_per_cat_knowledge} and Figure~\ref{fig:accuracy_per_cat_cognitive} assist in analyzing the following cases:

\begin{enumerate}
    \item accuracy between subcategories \label{enum:acc_sub}
    \item accuracy between subcategories and overall accuracy
    \item accuracy between different methods (AUC, F1) to define the perplexity threshold
    \item a comparison of all these between dataset flavors
\end{enumerate}

As mentioned in Section~\ref{sub:dataset_flavors}, answers that contain math and code do not behave in the same manner as plain text. Perplexity value is pretty low for both AI-generated and human written answers. An interesting behavior is induced by metacognitive category shown in Figure~\ref{fig:accuracy_per_cat_knowledge} and Figure~\ref{fig:accuracy_per_cat_cognitive}, which has surprisingly, the perfect accuracy score (100\% classification accuracy) and therefore needs further exploration. Metacognitive category, mostly includes answers that describe personal opinions and thoughts. LLMs at least till today, are not capable of expressing feelings, opinions and thoughts. Eventually, answers from students in this category have high perplexity values since the LLM model is not capable in producing convincing answers related to opinions and thoughts. To summarize all the above points and characteristics of the dataset we will focus on the original and !math !code flavors of the dataset.

In Figure~\ref{fig:accuracy_per_cat_knowledge} and Figure~\ref{fig:accuracy_per_cat_cognitive} the bars represent the classification accuracy if a text is AI-generated or not. Bars colored in purple and bars in green color represent the accuracy taking into consideration a perplexity threshold calculated by AUC and by F1 score methods accordingly. The dashed lines depict the classification accuracy if we do not leverage categories, with blue is accuracy using AUC method and red using F1 score method. In some cases the dashed lines overlap (same accuracy) and thus only one of them is displayed.

Starting with the knowledge dimension and the original dataset it can be observed that for both methods (AUC, F1) there is no gain at all for \textit{conceptual} and \textit{procedural} categories and marginal gain for \textit{factual} only for the F1 method, excluding the \textit{metacognitive} category. Moving on to the !math !code dataset flavor, there is significant gain in accuracy for all categories and for both threshold methods. To be more precise, there is a gain in accuracy of 4.32\%, 3.7\% and 8.1\% for \textit{conceptual}, \textit{factual} and \textit{procedural} knowledge dimensions respectively, using the AUC threshold method. For F1 score method the corresponding values are same except 9.74\% for the \textit{procedural} category. It can also be observed that \textit{procedural} category has better accuracy, 1.79\% when using AUC instead of F1 method, this though, is eliminated in the !math !code dataset flavor. The reason is that in the original dataset, we have bigger variations in the perplexity threshold (Table~\ref{tab:thr_per_knowledge_per_dataset}) compared to the !math !code flavor. Finally it can be observed that overall accuracy for both threshold methods is the same in the !math !code flavor, but differs slightly in the original dataset, with F1 method prevailing over AUC.

In the same way we can describe results shown in Figure~\ref{fig:accuracy_per_cat_cognitive}, which depicts the accuracy for each cognitive process dimension subcategory, but with one addition. These categories are not singular, meaning an answer maybe characterized by multiple values. This adds one more layer of complexity in the analysis. For the original dataset the categorization benefit is not obvious. Accuracy differs per method and per category a lot, with the only exception to be \textit{understand} and \textit{apply} categories. Accuracy with AUC is 5.26\% better in \textit{analyze} category compared to F1 method, the opposite for \textit{remember} category with 4.55\% worse results. For \textit{create} category, F1 method is better by 5.88\% from AUC. In the case of !math !code flavor, although these discrepancies are minimized, we have an unexpected behavior on the \textit{analyze} category and exactly the same results for \textit{remember}. In !math !code flavor, \textit{analyze} category has a drop of 25.77\% and 18.68\% for AUC and F1 methods accordingly, compared to the original dataset flavor. For the same flavor comparison, we have 28.3\% increase in accuracy for \textit{apply} category for both methods. The corresponding percentages for \textit{evaluate} and \textit{understand} are 11.11\% and decrease of 0.33\% accordingly. Finally \textit{create} category gained accuracy by 29.17\% for AUC and 25\% for F1 method. Again, we are not commenting on the other flavors as explained earlier.

Conclusively, the knowledge dimension indeed provides a better avenue for optimal threshold computation. This is reasoned by the experimental results depicted in this research study. Texts can be categorized and each category has different linguistic characteristics which eventually lead to different behavior of how humans write and AI generates the texts. Different perplexity thresholds essentially would better fit each category.

%% file: sections/6_discussion.tex
\subsection{Related Work}
Various tools exist in the internet like GPTZero\cite{GPTZero}, zeroGPT\cite{AITextDetector}, writer web app \cite{writer}, copyleaks \cite{copyleaks} even OpenAI foundation provided it's own classifier~\cite{AITextClassifier}. Most of them are free to use, but the methodology and implementation details are unknown. Our work, combined open-source frameworks and tools along with the knowledge from related literature, in order to evolve a methodology to identify text source, under the umbrella of a survey dataset~\cite{ibrahim2023perception}. There are also multiple recent research works \cite{tang2023science, mitchell2023detectgpt, munyer2023deeptextmark, li2023origin} on the source detection problem.

\subsection{Future Work}

This work focused on perplexity, as a hard threshold to classify text source given a limited dataset of human and ChatGPT generated answers to academic questions of various topics and domains. It is undeniable truth, that as LLMs are evolving the ability to distinguish the source would inevitably become extremely challenging, if not impossible, by using only a single metric.

Different linguistic metrics can be combined in order to create a text source signature that would fit the different profiles (human, AI). Our efforts will focus on identifying which are these linguistic metrics and how it would become possible to combine them. In parallel, more free models are becoming available to the public, thus, testing with different pre trained models and deciding which one captures more closely the behavior of ChatGPT would be feasible. Lastly, utilizing a more broad dataset would assist the better categorization, enabling the ability to classify texts and better match them against specialized class signatures.

A final aspect is to perform a deeper analysis on the cognitive dimension to better understand the dynamics created and if categorization on this level can be beneficial.

%% file: sections/7_conclusion.tex
In this work we used information theory perplexity metric in order to classify texts based on the entity that created them, either AI or human. This was enhanced by the usage of an academic dataset that also includes ChatGPT answers on questions in various academic domains and courses. The dataset was used in order to pre calculate perplexity scores for each answer and use these results to dynamically classify texts. Further analysis was done given a taxonomy on knowledge and cognitive process dimensions along with flag based attributes like `math` and `code`. We concluded that we have better results in the classification if we use the knowledge dimension categorization with adaptive perplexity thresholds per category and with excluding math and code related answers from the dataset.

%% file: bibs/references.bib
@online{openai-blog,
    title = {Introducing ChatGPT and Whisper APIs},
    url = {https://openai.com/blog/introducing-chatgpt-and-whisper-apis},
    urldate = {2023-05-01}
}

@online{openai-chatgpt,
    title = {OpenAI ChatGPT},
    url = {https://chat.openai.com/},
    urldate = {2023-05-01}
}

@online{openai-chatgpt-api,
    title = {OpenAI ChatGPT API},
    url = {https://platform.openai.com/docs/introduction},
    urldate = {2023-05-01}
}

@online{google-bard,
    title = {Google Bard},
    url = {https://bard.google.com/},
    urldate = {2023-05-01}
}

@online{gpt2-huggingface,
  title={GPT2-huggingface},
  url={https://huggingface.co/gpt2},
  urldate = {2023-05-01}
}

@article{radford2019language,
  title={Language models are unsupervised multitask learners},
  author={Radford, Alec and Wu, Jeffrey and Child, Rewon and Luan, David and Amodei, Dario and Sutskever, Ilya and others},
  journal={OpenAI blog},
  volume={1},
  number={8},
  pages={9},
  year={2019}
}

@online{gpt2-model,
    title = {GPT2 model},
    url = {https://huggingface.co/docs/transformers/main/en/model_doc/gpt2},
}

@online{GPT4,
    title = {GPT-4},
    url = {https://openai.com/product/gpt-4},
}

@online{LaMDA,
    title = {Google LaMDA},
    url = {https://blog.google/technology/ai/lamda/},
    urldate = {2023-05-01}
}

@article{adiwardana2020towards,
  title={Towards a Conversational Agent that Can Chat About… Anything},
  author={Adiwardana, Daniel and Luong, Thang},
  journal={Google AI Blog},
  year={2020}
}

@online{AI-Test-Kitchen,
    title = {AI Test Kitchen},
    url = {https://blog.google/technology/ai/join-us-in-the-ai-test-kitchen/},
}

@online{GPTZero,
    title = {GPTZero},
    url = {https://gptzero.me/},
}

@online{writer,
    title = {Writer},
    url = {https://writer.com/ai-content-detector/},
}

@online{copyleaks,
    title = {Copyleaks},
    url = {https://copyleaks.com/ai-content-detector},
}

@misc{AITextDetector, 
    month     = Jan,
    day       = 01,
    journal   = {AI Text Detector},
    publisher = {ZeroGPT},
    url       = {https://www.zerogpt.com},
    date      = 2023
}

@misc{AITextClassifier, 
    month     = Jan,
    day       = 31,
    journal   = {AI Text Classifier},
    publisher = {OpenAI},
    url       = {https://platform.openai.com/ai-text-classifier},
    date      = 2023,
}

@misc{ibrahim2023perception,
      title={Perception, performance, and detectability of conversational artificial intelligence across 32 university courses}, 
      author={Hazem Ibrahim and Fengyuan Liu and Rohail Asim and Balaraju Battu and Sidahmed Benabderrahmane and Bashar Alhafni and Wifag Adnan and Tuka Alhanai and Bedoor AlShebli and Riyadh Baghdadi and Jocelyn J. Bélanger and Elena Beretta and Kemal Celik and Moumena Chaqfeh and Mohammed F. Daqaq and Zaynab El Bernoussi and Daryl Fougnie and Borja Garcia de Soto and Alberto Gandolfi and Andras Gyorgy and Nizar Habash and J. Andrew Harris and Aaron Kaufman and Lefteris Kirousis and Korhan Kocak and Kangsan Lee and Seungah S. Lee and Samreen Malik and Michail Maniatakos and David Melcher and Azzam Mourad and Minsu Park and Mahmoud Rasras and Alicja Reuben and Dania Zantout and Nancy W. Gleason and Kinga Makovi and Talal Rahwan and Yasir Zaki},
      year={2023},
      eprint={2305.13934},
      archivePrefix={arXiv},
      primaryClass={cs.CY}
}

@misc{kant2018practical,
      title={Practical Text Classification With Large Pre-Trained Language Models}, 
      author={Neel Kant and Raul Puri and Nikolai Yakovenko and Bryan Catanzaro},
      year={2018},
      eprint={1812.01207},
      archivePrefix={arXiv},
      primaryClass={cs.CL}
}

@article{krathwohl2002revision,
  title={A revision of Bloom's taxonomy: An overview},
  author={Krathwohl, David R},
  journal={Theory into practice},
  volume={41},
  number={4},
  pages={212--218},
  year={2002},
  publisher={Taylor \& Francis}
}

@misc{gehrmann2019gltr,
      title={GLTR: Statistical Detection and Visualization of Generated Text}, 
      author={Sebastian Gehrmann and Hendrik Strobelt and Alexander M. Rush},
      year={2019},
      eprint={1906.04043},
      archivePrefix={arXiv},
      primaryClass={cs.CL}
}

@misc{hashimoto2019unifying,
      title={Unifying Human and Statistical Evaluation for Natural Language Generation}, 
      author={Tatsunori B. Hashimoto and Hugh Zhang and Percy Liang},
      year={2019},
      eprint={1904.02792},
      archivePrefix={arXiv},
      primaryClass={cs.CL}
}

@article{BROWN200624,
title = {Receiver operating characteristics curves and related decision measures: A tutorial},
journal = {Chemometrics and Intelligent Laboratory Systems},
volume = {80},
number = {1},
pages = {24-38},
year = {2006},
issn = {0169-7439},
doi = {https://doi.org/10.1016/j.chemolab.2005.05.004},
url = {https://www.sciencedirect.com/science/article/pii/S0169743905000766},
author = {Christopher D. Brown and Herbert T. Davis}
}

@misc{liu2023pretrain,
      title={Pre-train, Prompt and Recommendation: A Comprehensive Survey of Language Modelling Paradigm Adaptations in Recommender Systems}, 
      author={Peng Liu and Lemei Zhang and Jon Atle Gulla},
      year={2023},
      eprint={2302.03735},
      archivePrefix={arXiv},
      primaryClass={cs.IR}
}

@misc{vilar2022prompting,
      title={Prompting PaLM for Translation: Assessing Strategies and Performance}, 
      author={David Vilar and Markus Freitag and Colin Cherry and Jiaming Luo and Viresh Ratnakar and George Foster},
      year={2022},
      eprint={2211.09102},
      archivePrefix={arXiv},
      primaryClass={cs.CL}
}

@article{yadav2020sentiment,
  title={Sentiment analysis using deep learning architectures: a review},
  author={Yadav, Ashima and Vishwakarma, Dinesh Kumar},
  journal={Artificial Intelligence Review},
  volume={53},
  number={6},
  pages={4335--4385},
  year={2020},
  publisher={Springer}
}

@misc{ippolito2020automatic,
      title={Automatic Detection of Generated Text is Easiest when Humans are Fooled}, 
      author={Daphne Ippolito and Daniel Duckworth and Chris Callison-Burch and Douglas Eck},
      year={2020},
      eprint={1911.00650},
      archivePrefix={arXiv},
      primaryClass={cs.CL}
}

@misc{mitrović2023chatgpt,
      title={ChatGPT or Human? Detect and Explain. Explaining Decisions of Machine Learning Model for Detecting Short ChatGPT-generated Text}, 
      author={Sandra Mitrović and Davide Andreoletti and Omran Ayoub},
      year={2023},
      eprint={2301.13852},
      archivePrefix={arXiv},
      primaryClass={cs.CL}
}

@misc{sadasivan2023aigenerated,
      title={Can AI-Generated Text be Reliably Detected?}, 
      author={Vinu Sankar Sadasivan and Aounon Kumar and Sriram Balasubramanian and Wenxiao Wang and Soheil Feizi},
      year={2023},
      eprint={2303.11156},
      archivePrefix={arXiv},
      primaryClass={cs.CL}
}

@misc{celikyilmaz2021evaluation,
      title={Evaluation of Text Generation: A Survey}, 
      author={Asli Celikyilmaz and Elizabeth Clark and Jianfeng Gao},
      year={2021},
      eprint={2006.14799},
      archivePrefix={arXiv},
      primaryClass={cs.CL}
}

@misc{khalil2023chatgpt,
      title={Will ChatGPT get you caught? Rethinking of Plagiarism Detection}, 
      author={Mohammad Khalil and Erkan Er},
      year={2023},
      eprint={2302.04335},
      archivePrefix={arXiv},
      primaryClass={cs.AI}
}

@misc{gururangan2020dont,
      title={Don't Stop Pretraining: Adapt Language Models to Domains and Tasks}, 
      author={Suchin Gururangan and Ana Marasović and Swabha Swayamdipta and Kyle Lo and Iz Beltagy and Doug Downey and Noah A. Smith},
      year={2020},
      eprint={2004.10964},
      archivePrefix={arXiv},
      primaryClass={cs.CL}
}

@misc{tang2023science,
      title={The Science of Detecting LLM-Generated Texts}, 
      author={Ruixiang Tang and Yu-Neng Chuang and Xia Hu},
      year={2023},
      eprint={2303.07205},
      archivePrefix={arXiv},
      primaryClass={cs.CL}
}

@misc{mitchell2023detectgpt,
      title={DetectGPT: Zero-Shot Machine-Generated Text Detection using Probability Curvature}, 
      author={Eric Mitchell and Yoonho Lee and Alexander Khazatsky and Christopher D. Manning and Chelsea Finn},
      year={2023},
      eprint={2301.11305},
      archivePrefix={arXiv},
      primaryClass={cs.CL}
}

@misc{munyer2023deeptextmark,
      title={DeepTextMark: Deep Learning based Text Watermarking for Detection of Large Language Model Generated Text}, 
      author={Travis Munyer and Xin Zhong},
      year={2023},
      eprint={2305.05773},
      archivePrefix={arXiv},
      primaryClass={cs.MM}
}

@misc{li2023origin,
      title={Origin Tracing and Detecting of LLMs}, 
      author={Linyang Li and Pengyu Wang and Ke Ren and Tianxiang Sun and Xipeng Qiu},
      year={2023},
      eprint={2304.14072},
      archivePrefix={arXiv},
      primaryClass={cs.CL}
}
